\documentclass{article}
\usepackage{dblfloatfix}
\usepackage{graphicx}
\usepackage{soul}

\usepackage{booktabs} 
\usepackage{epstopdf}
\usepackage{comment}
\usepackage{tabularx}
\usepackage{subfigure}
\usepackage{paralist}
\usepackage{balance}
\usepackage{bm}
\usepackage{algorithm}
\usepackage{algorithmic}
\usepackage{ifthen}

\DeclareMathAlphabet\mathbfcal{OMS}{cmsy}{b}{n}

\usepackage{xcolor}
\definecolor{thedarkblue}{RGB}{0,0,120}  
\definecolor{mydarkblue}{rgb}{0,0.08,0.45}  
\definecolor{darkblue}{rgb}{0,0.08,180}
\colorlet{TufteRed}{red!80!black}

\definecolor{theblue}{RGB}{0,0,180}
\colorlet{thered}{TufteRed}

\usepackage{hyperref}
\hypersetup{%
    colorlinks=true,
    linkcolor=mydarkblue,
    citecolor=mydarkblue,
    filecolor=mydarkblue,
    urlcolor=mydarkblue}

\usepackage{microtype}
\usepackage{balance}

\usepackage{amsmath,amssymb,amsthm}

\usepackage{ragged2e}
\usepackage{multirow}
\usepackage{microtype}
\usepackage{balance}
\usepackage{setspace}

\graphicspath{{./}{./graphics/}}

\newcolumntype{R}[1]{>{\RaggedLeft\arraybackslash}} 
\newcolumntype{L}[1]{>{\RaggedRight\arraybackslash}}

\newcommand\TTT{\rule{0pt}{1.2ex}}
\newcommand\BBB{\rule[-1.0ex]{0pt}{0pt}}

\newcommand{\eg}{\emph{e.g.}}
\newcommand{\ie}{\emph{i.e.}}



\providecommand{\mat}[1]{\boldsymbol{\mathrm{#1}}}%
\renewcommand{\vec}[1]{\boldsymbol{\mathrm{#1}}}

\DeclareMathOperator{\hugeE}{\mbox{\huge\raise-0.3ex\hbox{E}}}
\DeclareMathOperator{\p}{\mathbb{P}}
\DeclareMathOperator{\hugep}{\mbox{\huge\raise-0.3ex\hbox{$\p$}}}

\newcommand{\RR}{\mathbb{R}}

\providecommand{\mA}{\ensuremath{\mat{A}}}

\providecommand{\mC}{\ensuremath{\mat{C}}}
\providecommand{\mD}{\ensuremath{\mat{D}}}

\providecommand{\mF}{\ensuremath{\mat{F}}}

\providecommand{\mH}{\ensuremath{\mat{H}}}
\providecommand{\mI}{\ensuremath{\mat{I}}}

\providecommand{\mL}{\ensuremath{\mat{L}}}

\providecommand{\mO}{\ensuremath{\mat{O}}}

\providecommand{\mR}{\ensuremath{\mat{R}}}
\providecommand{\mS}{\ensuremath{\mat{S}}}

\providecommand{\mU}{\ensuremath{\mat{U}}}

\providecommand{\mW}{\ensuremath{\mat{W}}}
\providecommand{\mX}{\ensuremath{\mat{X}}}
\providecommand{\mY}{\ensuremath{\mat{Y}}}

\providecommand{\mLambda}{\ensuremath{\mat{\Lambda}}}

\providecommand{\vb}{\ensuremath{\vec{b}}}

\providecommand{\vh}{\ensuremath{\vec{h}}}

\providecommand{\vw}{\ensuremath{\vec{w}}}
\providecommand{\vx}{\ensuremath{\vec{x}}}

\providecommand{\vz}{\ensuremath{\vec{z}}}

\providecommand{\vtheta}{\ensuremath{\vec{\theta}}}
\providecommand{\tmL}{\ensuremath{\tilde{\mat{L}}}}
\providecommand{\tmA}{\ensuremath{\tilde{\mat{A}}}}

\providecommand{\mPhi}{\ensuremath{\mathbf{\Phi}}}

\providecommand{\calX}{\ensuremath{\mathcal{X}}}

\providecommand{\calZ}{\ensuremath{\mathcal{Z}}}
\providecommand{\calR}{\ensuremath{\mathcal{R}}}

\newcommand{\Paragraph}[1]{\noindent\textbf{#1}}

\def\mTheta{\mathbf{\Theta}}
\def\mLambda{\mathbf{\Lambda}}

\newcommand{\gconv}{{\,\star_{\mathcal{G}}\,}}

\providecommand{\mR}{\ensuremath{\mat{R}}}

\usepackage{dsfont}
\usepackage{todonotes}

\usepackage{microtype}
\usepackage{graphicx}
\usepackage{subfigure}
\usepackage{booktabs}

\usepackage{hyperref}
\usepackage{cleveref}

\usepackage[accepted]{mlsys2020}

\usepackage{comment}

\mlsystitlerunning{Graph Deep Factors for Forecasting}

\begin{document}

\twocolumn[

\mlsystitle{Graph Deep Factors for Forecasting}

\mlsyssetsymbol{equal}{*}

\begin{mlsysauthorlist}
\mlsysauthor{Hongjie Chen}{vt}
\mlsysauthor{Ryan A. Rossi}{adobe}
\mlsysauthor{Kanak Mahadik}{adobe}
\mlsysauthor{Sungchul Kim}{adobe}
\mlsysauthor{Hoda Eldardiry}{vt}
\end{mlsysauthorlist}

\mlsysaffiliation{vt}{Department of Computer Science, Virginia Tech, Blacksburg, Virginia, USA}
\mlsysaffiliation{adobe}{Adobe Research, San Jose, California, USA}
\mlsyscorrespondingauthor{}{}

\mlsyskeywords{Graph Neural Network, Time-series Forecasting, Graph-based Time-series, Deep Probabilistic Forecasting, Deep Learning}

\vskip 0.3in

\begin{abstract}
Deep probabilistic forecasting techniques have recently been proposed for modeling large collections of time-series. However, these techniques explicitly assume either complete independence (local model) or complete dependence (global model) between time-series in the collection. This corresponds to the two extreme cases where every time-series is disconnected from every other time-series in the collection or likewise, that every time-series is related to every other time-series resulting in a completely connected graph. In this work, we propose a deep hybrid probabilistic graph-based forecasting framework called Graph Deep Factors (GraphDF) that goes beyond these two extremes by allowing nodes and their time-series to be connected to others in an arbitrary fashion. GraphDF is a hybrid forecasting framework that consists of a relational global and relational local model. In particular, we propose a relational global model that learns complex non-linear time-series patterns globally using the structure of the graph to improve both forecasting accuracy and computational efficiency. Similarly, instead of modeling every time-series independently, we learn a relational local model that not only considers its individual time-series but also the time-series of nodes that are connected in the graph. The experiments demonstrate the effectiveness of the proposed deep hybrid graph-based forecasting model compared to the state-of-the-art methods in terms of its forecasting accuracy, runtime, and scalability. Our case study reveals that GraphDF can successfully generate cloud usage forecasts and opportunistically schedule workloads to increase cloud cluster utilization by 47.5\% on average.
\end{abstract}
] 

\printAffiliationsAndNotice{}

\section{Introduction}
\label{sec:intro}
Cloud computing allows tenants to allocate and pay for resources on-demand. While convenient, this model requires users to provision the right amount of physical or \textit{virtualized} resources for every workload. The workloads often undergo striking variations in demand making static provisioning a poor fit. Moreover, allocating very few resources degrades the workload performance. 
Additionally, cloud platforms are exceptionally large and arduous to operate. Hence, optimizing their use is beneficial for both cloud operators and tenants. 
Accurate forecasting of workload patterns in cloud can aid in resource allocation, scheduling, and workload co-location decision for different workloads~\cite{clusterdata:Sliwko2018,clusterdata:Sebastio2018c,clusterdata:Sirbu2015}, massively reducing the operating costs and increasing resource efficiency~\cite{clusterdata:Liu2018gh}.
Learning must be fast to ensure precise decision-making and quick adaption to changes in demand. 

\begin{figure}[ht!]
    \centering
    \vspace{-7mm}
    \hspace{-4mm}
    \subfigure[Time-series of neighbors of a node in the graph]{
    \includegraphics[width=1.0\linewidth]{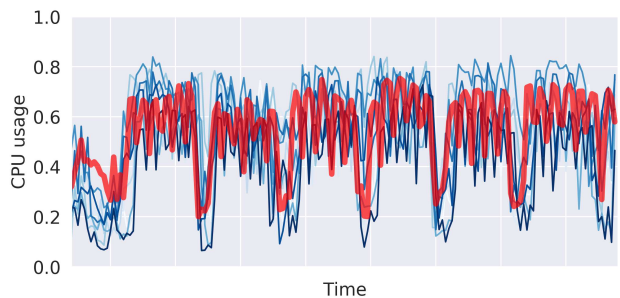}
    \label{fig:node-and-neighbors}
    }
    
    \vspace{-2mm}
    \hspace{-4mm}
    \subfigure[Time-series of randomly selected nodes]{
    \includegraphics[width=1.0\linewidth]{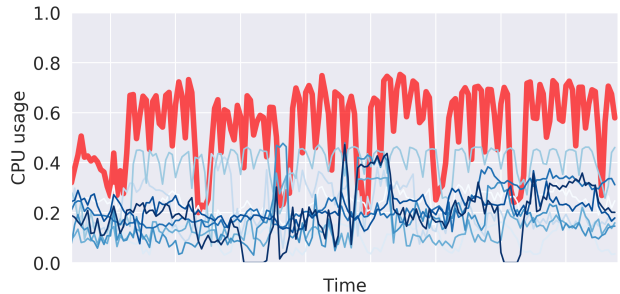}
    \label{fig:node-and-random-timeseries}
    }
    \vspace{-3mm}
    \caption{
    In (a) the time-series of CPU usage for a node (machine) in the Google workload data shown in \textcolor{red}{red} and its immediate neighbors in the graph (\textcolor{blue}{blue}) are highly correlated,
    whereas in (b) the time-series of randomly selected nodes are \emph{significantly different}. 
    }
    \label{fig:motivation}
    \vspace{-5mm}
\end{figure}

Previous work on time-series forecasting has focused mostly on local forecasting  models~\cite{brahim2004gaussian,girard2003gaussian} that treat each time-series independently or 
global forecasting models~\cite{Flunkert2017DeepARPF,wen2017multi,rangapuram2018deep} that consider all time-series jointly.
There has also recently been hybrid local-global models~\cite{wang2019deep,sen2019think} that attempt to combine the benefits of both~\cite{crawley2013r,gelman2013bayesian}. 
These past works all assume time-series are either completely independent or completely dependent.
However, these assumptions are often violated in practice as shown in Fig.~\ref{fig:motivation} where a node time-series of CPU usage from the Google workload data is shown to be 
highly dependent (correlated) on an arbitrary number of other node time-series.

In this work, we propose a deep hybrid \emph{graph-based} probabilistic forecasting model called Graph Deep Factors (GraphDF) that allows nodes and their time-series to be dependent (connected) in an arbitrary fashion.
GraphDF leverages a \emph{relational global model} that uses the dependencies between time-series in the graph to learn the complex non-linear patterns globally while leveraging a \emph{relational local model} to capture the individual random effects of each time-series locally.
The relational global model in GraphDF improves the runtime performance and scalability since instead of jointly modeling all time-series together (fully connected graph), which is computationally intensive, 
GraphDF learns the global latent factors that capture the complex non-linear time-series patterns among the time-series by leveraging only the graph that encodes the dependencies between the time-series.
GraphDF serves as a general framework for deep graph-based probabilistic forecasting as many components are completely interchangeable including the relational local and relational global models.

Relational local models in GraphDF use not only the individual time-series but also the neighboring time-series that are 1 or 2 hops away in the graph.
Thus, the proposed relational local models are more data efficient, especially when considering shorter time-series.
For instance, given an individual time-series with a short length (\eg, only 6 previous values), purely local models would have problems accurately estimating the parameters due to the lack of data points.
However, relational local models can better estimate such parameters by leveraging not only the individual time-series but the neighboring dependent time-series that are 1 or 2 hops away in the graph.

Relational global models in GraphDF are typically faster and more scalable since they avoid the pairwise dependence assumed by global models via the graph structure. 
By leveraging the dependencies between time-series encoded in the graph, GraphDF avoids a significant amount of work that would be required if the time-series are modeled jointly as done in Deep Factors (DF;~\citeauthor{wang2019deep}~\citeyear{wang2019deep}).

\subsection*{Main Contributions}
We propose a general and extensible deep hybrid graph-based probabilistic forecasting framework called Graph Deep Factors (GraphDF) that is capable of learning complex non-linear time-series patterns globally using the graph time-series data to improve both computational efficiency and forecasting accuracy while learning individual probabilistic models for individual time-series based on their own time-series and the collection of time-series from the immediate neighborhood of the node in the graph. The GraphDF framework is data-driven, fast, and scalable for applications requiring real-time performance such as
opportunistic scheduling in the cloud.

The state-of-the-art deep hybrid probabilistic forecasting methods focus on learning a global model that considers all time-series jointly as well as a local models learned from each individual time-series independently.
In this work, we propose a deep hybrid \emph{graph-based} probabilistic forecasting model that lies in between these two extremes. 
In particular, we propose a relational global model that learns complex non-linear time-series patterns globally using the structure of the graph to improve both computational efficiency and prediction accuracy. 
Similarly, instead of modeling every time-series independently, we learn a relational local model that not only considers its individual time-series but the time-series of nodes that are connected to an individual node in the graph.
Furthermore, GraphDF naturally generalizes many existing models including those based purely on local and global models, or a combination of both. This is due to its flexibility to interpolate between purely non-relational models (either local, global, or both) and relational models that leverage the graph structure encoding the dependencies between the different time-series.

The experiments demonstrate the effectiveness of the proposed deep hybrid graph-based probabilistic forecasting model in terms of its forecasting performance, runtime, and scalability.
Section~\ref{sec:opportunistic-scheduling} details a case study of applying GraphDF for forecasting cloud usage and scheduling batch workloads opportunistically to enhance utilization by 47.5\% on average.

\section{Related Work} \label{sec:related-work}
Classical methods such as ARIMA and exponential smoothing~\cite{ARIMA,Gardner1985ExponentialST} mainly target univariate forecasting tasks but fail to utilize relations between time-series.

Deep learning forecasting methods~\cite{ghaderi2017deep,lim2019enhancing,ahmed2010empirical,bai2018empirical,lim2018forecasting}, by contrast, are capable to perform multivariate forecasting by modeling the complex dependence among time-series.
The commonly used RNN structures~\cite{zhang2005neural,neural-forecasting-survey}, especially the variants LSTM~\cite{LSTM-1997,Bandara2020LSTMMSNetLF} and GRU~\cite{GRUchung2014}, excel at encoding both past information and current input for forecasting.
Further efforts have been made to accommodate various schemes and techniques, such as sequence-to-sequence models~\cite{bontempi2012machine,seq2seqNIPS2014,fan2019multi} for multi-step ahead prediction,
and attention mechanism~\cite{qin2017dual,DSANet} for dependency modeling.

While earlier work focuses on \textit{point forecasting} which aims at predicting optimal expected values, 
there is an increasing interest in \textit{probabilistic forecasting} models~\cite{wen2017multi,Alexandrov2019GluonTSPT,rangapuram2018deep,henaff2017prediction,maddix2018deep}.
Probabilistic models yield prediction as distributions and have the advantage of uncertainty estimates, which are important for downstream decision making.
Some recent probabilistic models are proposed in the multivariate manner, for example,
Salina et al. propose DeepAR~\cite{Flunkert2017DeepARPF}, a global model based on autoregressive recurrent networks and trained with all time-series in the same manner.
Wang et. al. propose DF~\cite{wang2019deep}, a hybrid global-local model that assumes time-series are determined by shared factors as well as individual randomness.
These methods indiscriminately model mutual dependence between time-series.
Hence, they imply a strong and unrealistic assumption that all time-series are pairwise related to one another in a uniformly equivalent way.

\Paragraph{Graph-based forecasting models}
Modeling the unique relations to each individual time-series from others naturally leads us to graph models~\cite{wang2019deepgraph},
among which Graph Neural Network(GNN)~\cite{GCN17} has recently showed great successes in extracting the information across nodes.
A tremendous amount of following work has been proposed to incorporate GNN with RNN~\cite{ST-MGCN,wang2018graph}, while most of them are limited in spatio-temporal study, such as traffic prediction~\cite{STGCN-ijcai2018} and ride-hailing demand forecasting~\cite{STDN-aaai2019,DMVST-aaai2018}.
Besides, these methods are not probabilistic models and they fail to deliver uncertainty estimates.

\Paragraph{Resource usage prediction}
Accurate forecasting is critical in resource usage prediction for a cloud platform to scale and schedule tasks according to the demand~\cite{Cloudplatform,venkataraman2014power}.
Existed work covers both traditional methods~\cite{workloadARIMA:QoS,zia2017adaptive}, machine learning approaches~\cite{crankshaw2017clipper} such as K-nearest neighbors~\cite{farahnakian2015utilization,shen2011cloudscale} and linear regression~\cite{6619533,yang2014cost} and RNN-based methods~\cite{RPPS,kumar2018workload,cloudpredict}.
However, none of these methods leverages a graph to model the relationships between nodes.

\section{Graph Deep Factors} \label{sec:framework}
In this section, we describe a general and extensible framework called Graph Deep Factors (GraphDF). It is capable of learning complex non-linear time-series patterns globally using the graph time-series data to improve both computational efficiency and performance while learning individual probabilistic models for individual time-series based on their own time-series and the collection of related time-series from the neighborhood of the node in the graph. 
The GraphDF framework is data-driven, flexible, accurate, and fast/scalable for large collections of multi-dimensional time-series data.

\subsection{Problem Formulation} \label{sec:problem-formulation}
We first introduce the deep graph-based probabilistic forecasting problem.
Notably, this is the first deep graph-based probabilistic forecasting framework. 
The framework is comprised of a relational graph global component (described in Section~\ref{sec:rel-global-model}) that learns the complex non-linear time-series patterns in the large collection of graph-based time-series data and
a relational local component (Section~\ref{sec:rel-local-component}) that handles uncertainty by learning a probabilistic forecasting model for every individual node in the graph that not only considers the time-series of the individual node, but also the time-series of nodes directly connected in the graph.

The proposed framework solves the following graph-based time-series forecasting problem.
Let $G=(V,E,\calX, \calZ)$ denote the graph model where $V$ is the set of nodes, $E$ is the set of edges, and $\calX=\{\mX^{(i)}\}_{i=1}^{N}$ is the set of covariate time-series associated with the $N$ nodes in $G$ where $\mX^{(i)} \in \RR^{D \times T}$ is the covariate time-series data associated with node $i$.
Hence, each node is associated with $D$ different covariate time-series.
Furthermore, $\calZ=\{\vz^{(i)}\}_{i=1}^{N}$ is the set of time-series associated with the $N$ nodes in $G$.
The $N$ nodes can be connected in an arbitrary fashion that reflects the dependence between nodes. 
Two nodes $i$ and $j$ that contain an edge $(i,j) \in E$ in the graph $G$ encodes an explicit dependency between the time-series data of node $i$ and $j$. 
Intuitively, using these explicit dependencies encoded in $G$ can lead to more accurate forecasts as shown in Fig,~\ref{fig:motivation}.

Further, let $\vz_{1:T}^{(i)}$ denote a univariate time-series for node $i$ in the graph where $\vz_{1:T}^{(i)} = \big[z_{1}^{(i)} \, \cdots \, z_{T}^{(i)}\big] \in \RR^{T}$ and $z_{t}^{(i)} \in \RR$. 
In addition, each node $i$ in the graph $G$ also has $D$ covariate time-series, $\mX^{(i)} \in \RR^{D \times T}$ where $\mX^{(i)}_{:,t} \in \RR^{D}$ (or $\vx^{(i)}_{t} \in \RR^{D}$) represents the $D$ covariate values at time step $t$ for node $i$.
We denote $\mA \in \RR^{N \times N}$ as the sparse adjacency matrix of the graph $G$ where $N=|V|$ is the number of nodes.
If $(i,j) \in E$, then $A_{ij}$ denotes the weight of the edge (dependency) between node $i$ and $j$.
Otherwise, $A_{ij}=0$ when $(i,j) \not\in E$.

We denote the unknown parameters in the model as $\mPhi$. 
Our goal is to learn a generative and probabilistic forecasting model described by $\mPhi$ that gives the (joint) distribution on target values in the future horizon $\tau$:
\begin{equation}\label{eq:problem}
\mathbb{P}\Big(\big\{\vz_{T+1:T+\tau}^{(i)}\big\}_{\!i=1}^{\!N} \,\Big|\,
\mA, \big\{\vz_{1 : T}^{(i)}, \mX_{:,1 : T+\tau}^{(i)}\!\big\}_{\!i=1}^{\!N}; \mPhi \Big)
\end{equation}\noindent
Hence, solving Eq.~\eqref{eq:problem} gives the joint probability distribution over future values given all covariates and past observations along with the graph structure represented by $\mA$ that encodes the explicit dependencies between the $N$ nodes and their corresponding time-series $\{\vz^{(i)}, \mX^{(i)}\!\}_{i=1}^{N}$.

\subsection{Framework Overview} \label{sec:framework-generative}
The Graph Deep Factors (GraphDF) framework aims to learn a parametric distribution to predict future workloads.
In GraphDF, each node $i$ and its time-series $z^{(i)}_{t}, \forall t=1,2,\ldots$ can be connected to other nodes and their time-series in an arbitrary fashion, which is encoded in the graph structure $G$. 
These connections represent explicit dependencies or correlations between the time-series of the nodes.
Furthermore, we also assume that each node $i$ and their time-series $\vz^{(i)}_{1:t}$ are governed by two key components including 
(1) a relational global model (Section~\ref{sec:rel-global-model}), and 
(2) a relational local random effect model (Section~\ref{sec:rel-local-component}).
As such, GraphDF is a hybrid forecasting framework.
Both the relational global component and relational local component of our framework leverage the graph and the way in which it is leveraged depends on the specific underlying model used for each component.

In the relational global component of GraphDF, we assume there are $K$ latent relational global factors that determine the fixed effect of each node and their time-series.
The relational global model consists of an approach that leverages the adjacency matrix $\mA$ of the graph $G$ and $\big\{\mX_{:,1:t}^{(j)}, \vz_{1:t-1}^{(j)}\big\}_{j=1}^{N}$ for learning the $K$ relational global factors that capture the relational non-linear time-series patterns in the graph-based time-series data,
\begin{equation} \label{eq:relationalg-final}
s_{k}(\cdot) = \textsc{gcrn}_k(\cdot), \quad k = 1,\ldots, K         
\end{equation}\noindent
where $s_{k}(\cdot),$\; $k=1,2,\ldots,K$ are the $K$ relational global factors that govern the underlying graph-based time-series data of all nodes in $G$.
In Eq.~\ref{eq:relationalg-final}, we learn the relational global factors using a Graph Convolutional Recurrent Network (GCRN)~\cite{gcrn17}, however, GraphDF is flexible for use with any other arbitrary deep time-series model, see Appendix~\ref{appendix:dcrnn} where we have also adapted DCRNN.
These are then used to obtain the relational global fixed effects function $c^{(i)}$ for node $i$ as follows,
\begin{equation} \label{eq:fixed-final}
c^{(i)}(\cdot) = \sum_{k=1}^K w_{i,k}\cdot s_{k}(\cdot) 
\end{equation}\noindent
where $w_{i,k}$ represents the $K$-dimensional embedding for node $i$.
Therefore, the final relational non-random fixed effect for node $i$ is simply a linear combination of the $K$ global factors and the embedding $\vw_i \in \RR^{K}$ for node $i$.
Now we use a relational local model discussed in Section~\ref{sec:rel-local-component} to obtain the local random effects for each node $i$.
More formally, we define the \emph{relational local random effects} function $b^{(i)}$ for a node $i$ in the graph $G$ as,
\begin{equation}\label{eq:random-final}
b^{(i)}(\cdot) \sim \mathcal{R}_i, \quad i = 1, \ldots, N
\end{equation}\noindent
where $\mathcal{R}_i$ can be any relational probabilistic time-series model. 
To compute $\mathbb{P}(\vz^{i}_{1:t}|\mathcal{R}_i)$ efficiently, we ensure $b^{(i)}_t$ obeys a normal distribution, and thus can be derived fast.
The \emph{relational latent function} of node $i$ denoted as $v^{(i)}$ is then defined as,
\begin{equation}\label{eq:vi-final} 
v^{(i)}(\cdot) = c^{(i)}(\cdot) + b^{(i)}(\cdot) 
\end{equation}\noindent
where $c^{(i)}$ is the relational fixed effect of node $i$ and $b^{(i)}$ is the relational local random effect for node $i$.
Hence, the relational latent function of node $i$ is simply a linear combination of the relational fixed effect $c^{(i)}$ from Eq.~\ref{eq:fixed-final} and its local relational random effect $b^{(i)}$ from Eq.~\ref{eq:random-final}.
Then, 
\begin{equation}\label{eq:emission-final}
z_{t}^{(i)} \sim \mathbb{P}\Big( z_{t}^{(i)} \, \big| \; v^{(i)}\big(\mA, \big\{\mX_{:,1:t}^{(j)}, \vz_{1:t-1}^{(j)}\big\}^{\!N}_{\!j=1}\big)\!\Big) 
\end{equation}\noindent
where the observation model $\mathbb{P}$ can be any parametric distribution.
For instance, $\mathbb{P}$ can be Gaussian, Poisson, Negative Binomial, among others.

The GraphDF framework is defined in
Eq.~\ref{eq:relationalg-final}-\ref{eq:emission-final}.
All the functions $s_k(\cdot), b^{(i)}(\cdot), v^{(i)}(\cdot)$
take past observations and covariates $\big\{\vz_{1 : t-1}^{(j)}, \mX_{:,1 : t}^{(j)}\!\big\}_{\!j=1}^{\!N}$, as well as the graph structure in the form of adjacency matrix $\mA$ as inputs.
We define $\vw_i = \big[w_{i,1} \cdots w_{i,k} \cdots w_{i,K}\big] \in \RR^{K}$ as the $K$-dimension embedding for time-series $\vz^{(i)}$ where $w_{i,k} \in \RR$ is the weight of the $k$-th factor for node $i$. 

\subsection{Relational Global Model}
\label{sec:rel-global-model}
The relational global model learns $K$ relational global factors from all time-series by a graph-based model.
These relational global factors are considered as the driving latent factors.
After the relational global factors are derived from the model, they are then used in a linear combination with weights given by embeddings for each time-series $\bm{w}_i$, as shown in Eq.~\eqref{eq:fixed-final}.

We first show how GCRN is modified for learning relational global factors in GraphDF.
Let $\vx_{t}^{(i)} \in \RR^{D}$ denote the $D$ covariates of node $i$ at time step $t$.
Now, we define the input temporal features of the relational global factor component with respect to graph $G$ as,
\begin{equation} \label{eq:Y_t_}
\mY_{t} = 
\begin{bmatrix}
z_{t-1}^{(1)} &  {\vx_{t}^{(1)}}^{\intercal} \\
\vdots & \vdots \\
z_{t-1}^{(N)} & {\vx_{t}^{(N)}}^{\intercal} 
\end{bmatrix} \in \RR^{N\times P}
\end{equation}\noindent
where $P=D+1$ for simplicity.
We refer to $\mY_t$ as a time-series graph signal.
The aggregation of information from other nodes is performed by a graph convolution operation
defined as the multiplication of a temporal graph signal with a filter $g_\theta$.
Given input features $\mY_t$, the graph convolution operation is denoted as $f_\gconv{\mTheta}$ with respect to the graph $G$ and parameters $\theta$:
\begin{align}
    f_\gconv{\mTheta}(\mY_t) &= g_\theta(\mL) \mY_t \label{eq:gcrn-graphconv} \\
    &= \mU g_\theta(\mLambda) \mU^T \mY_t \in \RR^{N\times P}
    \label{eq:gc}
\end{align}
where $\mL=\mI-\mD^{-\frac{1}{2}}\mA\mD^{-\frac{1}{2}}$ is the normalized Laplacian matrix of the adjacency matrix, 
$\mI \in \RR^{N\times N}$ is an identity matrix.
$D_{ii}=\sum_{j}A_{ij}$ is the diagonal weighted degree matrix.
$\mL=\mU \mathbf{\Lambda} \mU^T$ is the eigenvalue decomposition.
$g_\theta(\mLambda)=\text{diag}(\vtheta)$ denotes a filter parameterized by the coefficients $\vtheta \in \RR^{N}$ in the Fourier domain.
Directly applying Eq.~\eqref{eq:gc} is computationally expensive due to the matrix multiplication and the eigen-decomposition of $\mL$. 
To accelerate the computation, the Chebyshev polynomial approximation
up to a selected order $L-1$ is 
\begin{equation} \label{eq:filt_cheby}
	g_\theta(\mL) = \sum_{l=0}^{L-1} \theta_l T_l(\tmL),
\end{equation}
where $\vtheta = \big[\theta_0\,\cdots\,\theta_{L-1}\big] \in \RR^{L}$ in Eq.~\eqref{eq:filt_cheby} is the Chebyshev coefficients vector.
Importantly, $T_l(\tmL)=2\tmL T_{l-1}(\tmL) - T_{l-2}(\tmL)$ is recursively computed with the scaled Laplacian $\tmL=2\mL/\lambda_{\max}-\mI \in \RR^{N\times N}$, and starting values $T_0=1$ and $T_1=\tmL$.
The Chebyshev polynomial approximation improves the time complexity to linear in the number of edges $O(L|E|)$, \ie, number of dependencies between the multidimensional node time-series.
The order $L$ controls the local neighborhood time-series that are used for learning the relational global factors, \ie, a node's multi-dimensional time-series only depends on neighboring node time-series that are at maximum $L$ hops away in $G$.

Let $\mTheta \in \RR^{P \times Q \times L}$ be a tensor of parameters 
that maps the dimension $P$ of input to the dimension $Q$ of output:
\begin{align}
    \!\!\!\! \mH_{:, q} = \tanh\!\Bigg[\sum_{p=1}^{P} f_\gconv{\mTheta}(\mY_{t,\;:, p})\Bigg],\ \text{for} \; q \in {1\,\ldots\,Q}
\end{align}
The relational global component integrates the temporal dependence and relational dependence among nodes with the graph convolution,
\begin{align}
   \!\!\!\! &\mI_t = \sigma(\mTheta_{I} \gconv{}[\mY_t, \mH_{t-1}] + \mW_I\odot\mC_{t-1} + \vb_I) \\
\!\!\!\!    &\mF_t \!= \sigma(\mTheta_{F} \gconv{}[\mY_t, \mH_{t-1}] + \mW_F\odot\mC_{t-1} + \vb_F) \\
\!\!\!\!    &\mC_t \!=\! \mF_t\!\odot \mC_{t\text{-}1} \!+\! \mI_t \!\odot\! \tanh(\mTheta_{C} \!\gconv{}[\mY_t, \mH_{t\text{-}1}] + \!\vb_C) \\
\!\!\!\!    &\mO_t \!= \sigma(\mTheta_O\gconv{}[\mY_t, \mH_{t-1}] + \mW_O\odot \mC_t + \vb_O) \\
\!\!\!\!    &\mH_t \!= \mO_t \odot \tanh(\mC_t) \label{eq:gconvlstm-hid}
\end{align}\noindent
where $\mI_t\in \RR^{N\times Q}, \mF_t\in \RR^{N\times Q}, \mO_t \in \RR^{N\times Q}$ are the input, forget and output gate in the LSTM structure.
$Q$ is the number of hidden units,
$\mW_I\in \RR^{N\times Q}, \mW_F\in \RR^{N\times Q}, \mW_O \in \RR^{N\times Q}$ and $\vb_I, \vb_F, \vb_C, \vb_O \in \RR^{Q}$ are weights and bias parameters,
$\mTheta_I\in \RR^{P\times Q}, 
\mTheta_F \in \RR^{{P\times Q}}, 
\mTheta_C \in \RR^{{P\times Q}}, 
\mTheta_O \in \RR^{{P\times Q}}$ are parameters corresponding to different filters.

The hidden state $\mH_t \in \RR^{N\times Q}$ encodes the observation information from $\mH_{t-1}$ and $\mY_t$, as well as the relations across nodes through the graph convolution described by $\mTheta \gconv{}(\cdot)$ in Eq.~\eqref{eq:gcrn-graphconv}.
From hidden state $\mH_t$, we derive the value of $K$ relational global factors at time step $t$ as $\mS_t \in \RR^{N\times K} $ through a fully connected layer,
\begin{align}
    \mS_t = \mH_t \mW + \vb \label{eq:gcrn-st}
\end{align}
where $\mW \in \RR^{Q \times K}$ and $\vb \in \RR^{K}$ are the weight matrix and bias vector (for the $K$ relational global factors), respectively.
The relational global factors $\mS_t$ is derived from the Eq.~\eqref{eq:gcrn-st} that capture the complex non-linear time-series patterns between the different time-series globally.

Finally, the fixed effect at time $t$ is derived for each node $i$ as a weighted sum with the embedding $\vw_{i} \in \RR^{K}$ and the relational global factors $\mS_t$, as
\begin{equation}
    c_{t}^{(i)}(\cdot) = \sum_{k=1}^K w_{i,k} \cdot S_{i,k,t} \label{eq:gcrn-cit}
\end{equation}
The embedding $\vw_i$ represents the weighted contribution that each relational factor has on node $i$.

\subsection{Relational Local Model}\label{sec:rel-local-component}
The (stochastic) relational local component handles uncertainty by learning a probabilistic forecasting model for every individual node in the graph that not only considers the time-series of the individual node, but also the time-series of nodes directly connected in the graph. 
This has the advantage of improving both forecasting accuracy and data efficiency.
GraphDF is therefore able to make more accurate forecasts further in the future with less training data.

The random effects in the relational local model represent the local fluctuations of the individual node time-series.
The relational local random effect for each node time-series $b^{(i)}$ is sampled from the relational local model $\calR_{i}$, as shown in Eq.~\eqref{eq:random-final}.
For $\calR_i$, we choose the Gaussian distribution as the likelihood function for sampling, but other parametric distributions such as Student-t or Gamma distributions are also possible. 
Compared to the relational global component of GraphDF from Section~\ref{sec:rel-global-model} that uses the entire graph $G$ along with all the node multi-dimensional time-series to learn $K$ global factors that capture the most important non-linear time-series patterns in the graph-based time-series data, the relational local component focuses on modeling an individual node $i \in V$ and therefore leverages only the time-series of node $i$ and the set of highly correlated time-series from its immediate local neighborhood $\Gamma_i$.
Hence, $\{\vz^{(j)}, \mX^{(j)}\}, j \in \Gamma_i$.
Intuitively, the relational local component of GraphDF achieves better data efficiency by leveraging the highly correlated neighboring time-series along with its own time-series.
This allows GraphDF to make more accurate forecasts further in the future with less training data.
We now introduce probabilistic GCRN 
that can be used as the stochastic relational local component in GraphDF.

\begin{table*}[ht!]
\vspace{-4mm}
	\centering
	\caption{Dataset statistics and properties}
	\label{table:dataset-statistics}
	\vspace{1mm}
	\small
	\footnotesize
	\begin{tabular}{l rr cccccccc}
		\toprule
		 &  & &  & Avg. & Median & Mean &  & & Median \\ 
		Data & $|V|$ & $|E|$ & Density & Deg. & Deg. & wDeg. & Time-scale & T usage & CPU usage\\ 
		\midrule
\TTT\BBB
		\textbf{Google} & 12,580 & 1,196,658 & 0.0075 & 95.1 & 40 & 30.3 & 5 min & 8,354 &  21.4\% \\

		\textbf{Adobe} & 3,270 & 221,984 & 0.0207 & 67.9 & 15 & 67.7 & 30 min & 1,687 & 9.1\% \\
		\bottomrule
	\end{tabular}
	\vspace{-2mm}
\end{table*}


In contrast to the relational global model in Section~\ref{sec:rel-global-model}, the relational local model focuses on learning an individual local model for each individual node based on its own multi-dimensional time-series data as well as the nodes neighboring it.
This enables us to model the local fluctuations of the individual multi-dimensional time-series data of each node.
Compared to RNN, the benefits of the proposed probabilistic GCRN model in the local component is that it not only models the sequential nature of the data, but also exploits the graph structure by using the surrounding nodes to learn a more accurate model for each individual node in $G$.
This is an ideal property for we assume the fluctuations of each node are related to those of other connected nodes in the $\ell$-localized neighborhood, which was shown to be the case in Fig.~\ref{fig:motivation}.

Let $C = \Gamma_i$ denote the set of neighbors of a node $i$ in the graph $G$.
Note that $C$ can be thought of as the set of related neighbors of node $i$, which may be the immediate 1-hop neighbors, or more generally, the $\ell$-hop neighbors of $i$.
Recall that we define $\vx_{t}^{(i)} \in \RR^{D}$ as the $D$ covariates of node $i$ at time $t$.
Then, we define $\mX_t^{C}$ as an $|C| \times D$ matrix consisting of the covariates of all the neighboring nodes $j \in C$ of node $i$.
Now, we define the input temporal features of the relational local model for node $i$ as,
\begin{equation} \label{eq:Y_t_rel-local}
\mY_{t}^{(i)} = 
\begin{bmatrix}
z_{t-1}^{(i)} &  {\vx_{t}^{(i)}}^{\intercal} \\
\vz_{t-1}^{(C)} &  \mX_{t}^{(C)}
\end{bmatrix}
\end{equation}\noindent
Let $\mL^{(i)} \in \RR^{(|C|+1)\times(|C|+1)}$ denote the submatrix of Laplacian matrix $\mL$ that consist of rows and columns corresponding to node $i$ and its neighbors $C$.
For each node $i$, we derive the relational local random effect using its past observations and covariates of the node $i$ and those of its neighbors through the graph convolution regarding $\mL^{(i)}$.
\begin{align} \nonumber
  &  \mI_t^{(i)} \!= \sigma(\mTheta_{I}^{(i)} \gconv{}[\mY_t^{(i)}, \; \mH_{t-1}^{(i)}] + \mW_I^{(i)}\odot\mC_{t-1}^{(i)} + \vb_I^{(i)})\\ \nonumber
&    \mF_t^{(i)} \!= \sigma(\mTheta_{F}^{(i)} \gconv{}[\mY_t^{(i)}, \; \mH_{t-1}^{(i)}] + \mW_F^{(i)}\odot\mC_{t-1}^{(i)} + \vb_F^{(i)})\\ \nonumber
   & \mC_t^{(i)} \!= \!\mF_t^{(i)}\!\odot\! \mC_{t-1}^{(i)} \!+ \!\mI_t^{(i)} \!\odot\! \tanh(\mTheta_{C}^{(i)} \gconv{}[\mY_t^{(i)}, \mH_{t-1}^{(i)}] + \vb_C^{(i)}) \\ \nonumber
    &\mO_t^{(i)} \!= \sigma(\mTheta_{O}^{(i)} \gconv{}[\mY_t^{(i)}, \; \mH_{t-1}^{(i)}] + \mW_O^{(i)}\odot \mC_t^{(i)} + \vb_O^{(i)}) \\ \nonumber
   & \mH_t^{(i)} \!= \mO_t^{(i)} \odot \tanh(\mC_t^{(i)}) \label{eq:local-gconvlstm-hid}
\end{align}
where 
$\mTheta_{I}^{(i)} \in \RR^{P\times R}$, $\mTheta_{F}^{(i)} \in \RR^{P\times R}$, $\mTheta_{C}^{(i)} \in \RR^{P\times R}$, $\mTheta_{O}^{(i)} \in \RR^{P\times R}$ denote the parameters corresponding to different filters of the relational local model,
$R$ is the number of hidden units in the relational local model, and recall $P=D+1$.
Further, $\mH_t^{(i)} \in \RR^{(|C|+1)\times R}$ is the hidden state for node $i$ and its neighbors $\Gamma_i$.
$\mW_I^{(i)}\in\RR^{(|C|+1)\times R},
\mW_F^{(i)}\in\RR^{(|C|+1)\times R},
\mW_O^{(i)}\in\RR^{(|C|+1)\times R}$ are weight matrix parameters and 
$\vb_{I}^{(i)}\in \RR^{R}, 
\vb_{F}^{(i)}\in \RR^{R}, 
\vb_{C}^{(i)}\in \RR^{R}, 
\vb_{O}^{(i)}\in \RR^{R}$ are bias vector parameters.
Note in the above formulation, we assume $\ell=1$, hence, only the immediate 1-hop neighbors are used.

From the hidden state $\mH_t^{(i)}$, we only take the row corresponding to node $i$ to derive the relational local random effect for node $i$.
We denote the value as $\vh_t^{(i)} \in \RR^{R}$,
and apply a fully connected layer with a softplus activation function to aggregate the hidden units,
\begin{equation}
    \sigma_{i,t} = 
    \log
    \big(
    \exp({\vw^{(i)}}^{\intercal}\vh_t^{(i)} + \beta^{(i)})+1
    \big) \label{eq:sigma-gcrn}
\end{equation}
where $\vw^{(i)} \in \RR^{R}$ and $\beta^{(i)}$ are weight vector and bias, respectively.

Finally, the relational local random effect $b_t^{(i)}(\cdot)$ for node $i$ at time $t$ is sampled from a Gaussian distribution with zero mean and a variance given by $\sigma^2$ in Eq.\eqref{eq:sigma-gcrn},
\begin{align}
    b_{t}^{(i)}(\cdot) &\sim \mathcal{N}(0, \sigma_{i,t}^2) \label{eq:gcrn-random}
\end{align}
The relational local random effect $b_t^{(i)}$ captures both past observations, covariate values of node $i$ and its neighbors $\Gamma_i$ for uncertainty estimates through $\sigma_{i,t}$.
A small $\sigma_{i,t}$ means a low uncertainty of prediction for node $i$ at $t$.
Specifically, the probabilistic model subsumes the point forecasting model when the relational local random effect is zero for all nodes at all time steps as $\sigma_{i,t}=0, \forall i \forall t$.
The probabilistic property also allows the uncertainty to be propagated forward in time.

\begin{table*}[!t]
\small
\centering
\setlength{\tabcolsep}{4.6pt}
\caption{Results for one-step ahead forecasting (\textsc{p50ql} and \textsc{p90ql}).
}
\vspace{1mm}
\label{table:results-one-step-p50ql-p90ql}
\begin{tabular}{
@{}llccccccc
}
\toprule
& \multirow{1}{*}{\textsc{data}} & NBEATS & MQRNN & DeepAR & DF   & GraphDF-GG & GraphDF-GR & GraphDF-RG \\
\midrule
\TTT\BBB
\multirow{2}{*}{\textsc{p50ql}}  &
\multirow{1}{*}{\textbf{Google}} &
10.064\ \!$\pm$\ \!62.117        & 0.172\ \!$\pm$\ \!0.001 & 0.098\ \!$\pm$\ \!0.001 & 0.239\ \!$\pm$\ \!0.001 & \textbf{0.072\ \!$\pm$\ \!0.000} & 0.076\ \!$\pm$\ \!0.000 & 0.077\ \!$\pm$\ \!0.000 \\
 & \multirow{1}{*}{\textbf{Adobe}} &
3.070\ \!$\pm$\ \!2.286          & 0.272\ \!$\pm$\ \!0.001 & 0.619\ \!$\pm$\ \!0.026 & 1.649\ \!$\pm$\ \!0.001 & \textbf{0.188\ \!$\pm$\ \!0.000} & 0.210\ \!$\pm$\ \!0.001 & 0.746\ \!$\pm$\ \!0.835 \\

\midrule
\TTT\BBB
\multirow{2}{*}{\textsc{p90ql}}  &
\multirow{1}{*}{\textbf{Google}} &
2.013\ \!$\pm$\ \!2.485          & 0.106\ \!$\pm$\ \!0.001 & 0.051\ \!$\pm$\ \!0.000 & 0.144\ \!$\pm$\ \!0.002 & \textbf{0.041\ \!$\pm$\ \!0.000} & 0.044\ \!$\pm$\ \!0.000 & 0.048\ \!$\pm$\ \!0.000 \\
& \multirow{1}{*}{\textbf{Adobe}} &
5.524\ \!$\pm$\ \!7.410          & 0.217\ \!$\pm$\ \!0.000 & 0.949\ \!$\pm$\ \!0.086 & 1.802\ \!$\pm$\ \!0.002 & \textbf{0.153\ \!$\pm$\ \!0.001} & 0.169\ \!$\pm$\ \!0.001 & 0.342\ \!$\pm$\ \!0.037 \\
\bottomrule
\end{tabular}
\end{table*}

\subsection{Learning \& Inference} \label{sec:framework-learning}
To train a GraphDF model, 
we estimate the parameters $\mPhi$, which represent all trainable parameters ($\mW$, etc.) in the relational global and relational local model, as well as the parameters in the embeddings.
We leverage the maximum likelihood estimation,
\begin{equation}
\label{eq:parameters-optimization}
\mPhi = \text{argmax} \sum_i \mathbb{P}\big(\vz^{(i)} \big| \mPhi, \mA, \big\{\mX_{:,1:t}^{(j)}, \vz_{1:t-1}^{(j)}\big\}^{\!N}_{\!j=1} \!\big)
\end{equation}
where 
\begin{equation}
\label{eq:neg-log-like}
\mathbb{P}(\vz^{(i)}) = \sum_{t}-\frac{1}{2}\ln(2\pi\sigma_{i,t})- \sum_{t}\frac{(z_{t}^{(i)} - c_{i,t})^2}{2\sigma_{i,t}^2}
\end{equation}
is the negative log likelihood of Gaussian function.
Notice that maximizing $-\frac{1}{2}\ln(2\pi\sigma_{i,t})$ will minimize the relational local random effect,
at the same time, $\sigma$ is small when the predicted fixed effect $c_{i,t}$ is close to the actual value $z_t^{(i)}$, as shown in the second term $\frac{(z_t^{(i)}-c_{i,t})}{2\sigma_{i,t}^2}$ in Eq.~\eqref{eq:neg-log-like}.
We provide a general training approach in the Appendix~\ref{appendix-opportunistic-learning}(see Algorithm~\ref{alg:graphDF-training} for details).

\subsection{Model Variants} \label{sec:model-variants}
In this section, we define a few of the GraphDF model variants investigated in Section~\ref{sec:exp}.
\begin{itemize}
    \item \textbf{GraphDF-GG:} This is the default model in our GraphDF framework, where we use a graph model to learn the $K$ relational global factors (Sec.~\ref{sec:rel-global-model}) and the probabilistic local graph component from Sec.~\ref{sec:rel-local-component} as the relational local model.
        
    \item \textbf{GraphDF-GR:} This model variant from the GraphDF framework uses the GCRN from Sec.~\ref{sec:rel-global-model} to learn the $K$ relational global factors from the graph-based time-series data and leverages a simple RNN for modeling the local random effects of each node.
    \item \textbf{GraphDF-RG:} This GraphDF model variant uses a simple RNN to learn the $K$ global factors 
    and the probabilistic graph component from Sec.~\ref{sec:rel-local-component} as the relational local model.
\end{itemize}

The GraphDF framework is flexible with many interchangeable components including 
the relational global component (Sec.~\ref{sec:rel-global-model}) that uses the graph-based time-series data to learn the $K$ global factors and fixed effects of the nodes as well as 
the relational local model (Sec.~\ref{sec:rel-local-component}) for obtaining the relational local random effects of the nodes.

\section{Experiments} \label{sec:exp}
The experiments are designed to investigate the following:
(RQ1) Does GraphDF outperform previous state-of-the-art deep probabilistic forecasting methods?
(RQ2) Are the GraphDF models fast and scalable for large-scale time-series forecasting?
(RQ3) Can GraphDF generate cloud usage forecasts to effectively perform opportunistic workload scheduling?

\subsection{Experimental setup}
We used two real-world cloud traces from Google~\cite{reiss2011google} and Adobe. The statistics and properties of these two datasets are in Table~\ref{table:dataset-statistics}.
The Google dataset consists of 167 Gigabytes records of machine resource usage.
Further details are described in the Appendix~\ref{appendix-data-details}.

For the opportunistic workload scheduling case study later in Section~\ref{sec:opportunistic-scheduling}, we need to train a model \emph{fast} within minutes and then forecast a single as well as multiple time steps ahead, which are then used to make opportunistic scheduling and scaling decisions.
Therefore, to ensure the models are trained fast within minutes, we use only the most recent $6$ observations in the time-series data for training.
We set the number of embedding dimension as $K=10$ in $\bm{w}_i \in \RR^{K}$ and use time feature series as covariates.
We set the embedding dimension to $K=10$ in $\bm{w}_i \in \RR^{K}$ and used $D=5$ covariates
for each time-series. 
Similar to DF~\cite{wang2019deep}, the time features (\eg minute of hour, hour of day) are used as covariates. We derive a fixed graph using Radial Based Function (RBF) on the past observations.
See Appendix~\ref{appendix-data-details} for further details.

\begin{table*}[!t]
\vspace{-2mm}
\setlength{\tabcolsep}{5.2pt} 
\centering
\caption{Results for multi-step ahead forecasting (\textsc{p50ql}).
}
\vspace{1mm}
\label{table:results-multi-step-ahead-forecasting-p50ql}
\small
\begin{tabular}{@{}l@{} cccccccc 
}
\toprule
\multirow{1}{*}{\textsc{data}} 
& \multirow{1}{*}{\textsc{h}} & NBEATS                  & MQRNN                   & DeepAR                  & DF                      & GraphDF-GG                       & GraphDF-GR              & GraphDF-RG \\
\midrule
\multirow{3}{*}{\textbf{Google}\;\;}
& 3 & 0.741\ \!$\pm$\ \!0.050 & 0.257\ \!$\pm$\ \!0.011 & 0.148\ \!$\pm$\ \!0.001 & 0.400\ \!$\pm$\ \!0.004 & \textbf{0.091\ \!$\pm$\ \!0.001} & 0.134\ \!$\pm$\ \!0.002 & 0.097\ \!$\pm$\ \!0.000 \\
& 4 & 0.618\ \!$\pm$\ \!0.105 & 0.410\ \!$\pm$\ \!0.017 & 0.191\ \!$\pm$\ \!0.000 & 0.454\ \!$\pm$\ \!0.007 & \textbf{0.097\ \!$\pm$\ \!0.002} & 0.185\ \!$\pm$\ \!0.002 & 0.109\ \!$\pm$\ \!0.000 \\
& 5 & 0.485\ \!$\pm$\ \!0.021 & 0.684\ \!$\pm$\ \!0.012 & 0.466\ \!$\pm$\ \!0.006 & 0.563\ \!$\pm$\ \!0.017 & 0.128\ \!$\pm$\ \!0.000          & 0.284\ \!$\pm$\ \!0.012 & \textbf{0.126\ \!$\pm$\ \!0.001} \\

\midrule
\multirow{3}{*}{\textbf{Adobe}}
& 3 & 1.683\ \!$\pm$\ \!0.100 & 0.556\ \!$\pm$\ \!0.028 & 0.592\ \!$\pm$\ \!0.017 & 1.116\ \!$\pm$\ \!0.006 & \textbf{0.272\ \!$\pm$\ \!0.004} & 0.315\ \!$\pm$\ \!0.006 & 0.319\ \!$\pm$\ \!0.005 \\
& 4 & 1.424\ \!$\pm$\ \!0.210 & 0.574\ \!$\pm$\ \!0.011 & 0.629\ \!$\pm$\ \!0.024 & 1.029\ \!$\pm$\ \!0.001 & \textbf{0.314\ \!$\pm$\ \!0.004} & 0.353\ \!$\pm$\ \!0.007 & 0.405\ \!$\pm$\ \!0.007 \\
& 5 & 1.069\ \!$\pm$\ \!0.027 & 0.687\ \!$\pm$\ \!0.064 & 0.633\ \!$\pm$\ \!0.012 & 1.039\ \!$\pm$\ \!0.004 & \textbf{0.375\ \!$\pm$\ \!0.007} & 0.401\ \!$\pm$\ \!0.014 & 0.484\ \!$\pm$\ \!0.005 \\

\bottomrule
\end{tabular}
\vspace{-2mm}
\end{table*}
\begin{table*}[!t]
\vspace{-2mm}
\setlength{\tabcolsep}{5.2pt} 
\centering
\caption{Results for multi-step ahead forecasting (\textsc{p90ql}).
}
\vspace{1mm}
\label{table:results-multi-step-ahead-forecasting-p90ql}
\small
\begin{tabular}{@{}l@{} cccccccc 
}
\toprule
\multirow{1}{*}{\textsc{data}} 
& \multirow{1}{*}{\textsc{h}} & NBEATS & MQRNN & DeepAR & DF & GraphDF-GG & GraphDF-GR & GraphDF-RG \\
\midrule
\multirow{3}{*}{\textbf{Google}\;\;}
& 3 & 0.830\ \!$\pm$\ \!0.262 & 0.091\ \!$\pm$\ \!0.001 & 0.067\ \!$\pm$\ \!0.000 & 0.208\ \!$\pm$\ \!0.002 & \textbf{0.051\ \!$\pm$\ \!0.000} & 0.051\ \!$\pm$\ \!0.000 & 0.089\ \!$\pm$\ \!0.000 \\
& 4 & 0.976\ \!$\pm$\ \!0.429 & 0.090\ \!$\pm$\ \!0.000 & 0.070\ \!$\pm$\ \!0.000 & 0.213\ \!$\pm$\ \!0.002 & \textbf{0.050\ \!$\pm$\ \!0.001} & 0.076\ \!$\pm$\ \!0.001 & 0.095\ \!$\pm$\ \!0.001 \\
& 5 & 0.523\ \!$\pm$\ \!0.085 & 0.124\ \!$\pm$\ \!0.000 & 0.134\ \!$\pm$\ \!0.000 & 0.220\ \!$\pm$\ \!0.002 & \textbf{0.069\ \!$\pm$\ \!0.001} & 0.167\ \!$\pm$\ \!0.013 & 0.094\ \!$\pm$\ \!0.001 \\

\midrule
\multirow{3}{*}{\textbf{Adobe}}
& 3 & 2.556\ \!$\pm$\ \!0.328 & 0.317\ \!$\pm$\ \!0.002 & 0.751\ \!$\pm$\ \!0.117 & 1.545\ \!$\pm$\ \!0.008 & \textbf{0.248\ \!$\pm$\ \!0.004} & 0.254\ \!$\pm$\ \!0.003 & 0.301\ \!$\pm$\ \!0.003 \\
& 4 & 1.862\ \!$\pm$\ \!1.212 & 0.335\ \!$\pm$\ \!0.004 & 0.696\ \!$\pm$\ \!0.170 & 1.673\ \!$\pm$\ \!0.008 & \textbf{0.317\ \!$\pm$\ \!0.006} & 0.318\ \!$\pm$\ \!0.009 & 0.482\ \!$\pm$\ \!0.014 \\
& 5 & 1.512\ \!$\pm$\ \!0.082 & 0.463\ \!$\pm$\ \!0.003 & 0.546\ \!$\pm$\ \!0.000 & 1.690\ \!$\pm$\ \!0.015 & \textbf{0.410\ \!$\pm$\ \!0.021} & 0.434\ \!$\pm$\ \!0.007 & 0.512\ \!$\pm$\ \!0.007 \\

\bottomrule
\end{tabular}
\vspace{-1mm}
\end{table*}


\subsection{Forecasting Performance} \label{sec:exp-forecasting-performance}
We investigate the proposed GraphDF framework with various horizons including $\tau = \{1,3,4,5\}$. 
We evaluate the three GraphDF variants described in Section~\ref{sec:model-variants} against 
four state-of-the-art probabilistic forecasting methods including Deep Factors, \text{DeepAR}~\cite{Flunkert2017DeepARPF}, \text{MQRNN}~\cite{wen2017multi}, and \text{NBEATS}~\cite{oreshkin2019n}.
Deep Factors is a generative approach that combines a global model and a local model.
In DF, we use the Gaussian likelihood in terms of the random effects in the deep factors model.
We use 10 global factors with a LSTM cell of 1-layer and 50 hidden units in its global component, and 1-layer and 5 hidden units RNN in the local component.
\text{DeepAR} is an RNN-based global model, we use a LSTM layer with 50 hidden units in DeepAR.
\text{MQRNN} is a sequence model with quantile regression and \text{NBEATS} is an interpretable pure deep learning model.
For MQRNN, we use a GRU bidirectional layer with 50 hidden units as encoder and a modified forking layer in decoder.
For N-BEATS, we use an ensemble modification of the model and take the median value from 10 bagging bases as results.
All methods are implemented using MXNet Gluon~\cite{chen2015mxnet,Alexandrov2019GluonTSPT}.
The Adam optimization method is used with an initial learning rate as 0.001 to train all models.
The training epochs are selected by grid search in $\{100,200,\ldots,1000\}$. 
An early stopping strategy is leveraged if weight losses do not decrease for 10 continuous epochs.
Details on the hyperparameter tuning for each method
are provided in Appendix~\ref{appendix:hyperparameter-tuning}.
To evaluate the probabilistic forecasts, we use $\rho$-quantile loss~\cite{wang2019deep}.
We run 10 trials and report the average 
for $\rho=\{0.1, 0.5, 0.9\}$, denoted as the P10QL, P50QL and P90QL, respectively.
Further details on $\rho$-quantile loss are provided in Appendix~\ref{appendix:prob-forecast-eval-metric}.

The results for single and multi-step ahead forecasting are provided in Table~\ref{table:results-one-step-p50ql-p90ql} and Table~\ref{table:results-multi-step-ahead-forecasting-p50ql}-\ref{table:results-multi-step-ahead-forecasting-p90ql}, respectively, where the best result for every dataset and forecast horizon are highlighted in bold.
In all cases, we observe that the GraphDF models outperform previous state-of-the-art methods across all datasets and forecast horizons. 
Furthermore, the GraphDF-GG variant outperforms the other variants in nearly all cases.
Additional results have been removed for brevity. However, we also observed similar performance for P10QL, (Table~\ref{table:results-one-step-p10ql}-\ref{table:results-multi-step-ahead-forecasting-p10ql} in the Appendix~\ref{appendix-additional-results}.)
%
\begin{figure}[!th]
\vspace{-2mm}
\centering
\includegraphics[width=0.9\linewidth]{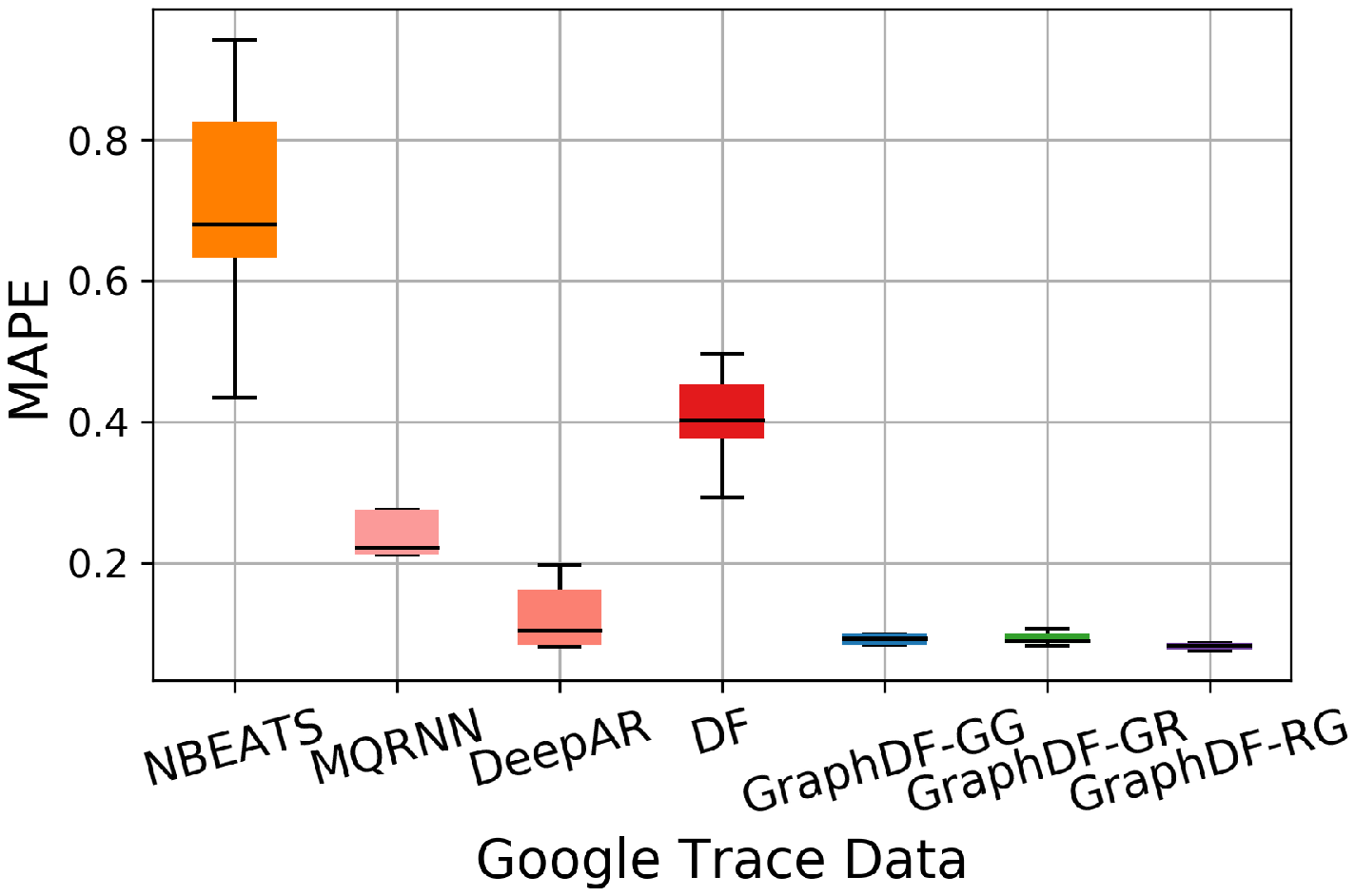}
\includegraphics[width=0.9\linewidth]{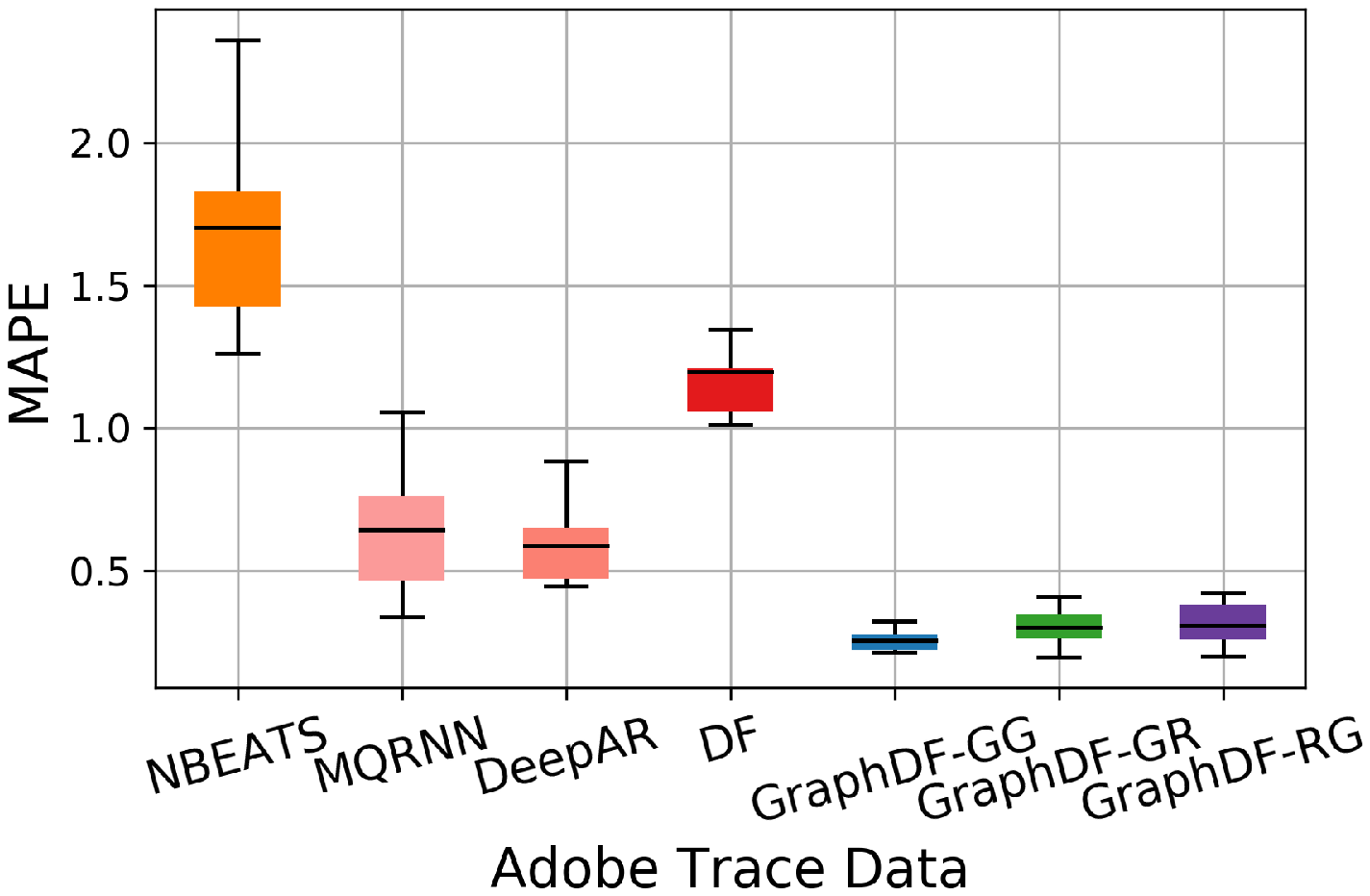}
\vspace{-3mm}
\caption{
Probabilistic forecasting results for 3-step ahead forecast horizon (P50QL, or equivalently, MAPE).
Overall, GraphDF and its variants 
significantly outperform the other methods 
while having much lower variance.
}
\label{fig:results-boxplots-p50ql}
\vspace{-1mm}
\end{figure}
%
To understand the overall performance and variance of the results, we show boxplots for each model in Fig.~\ref{fig:results-boxplots-p50ql}.
Strikingly, we observe that the GraphDF models provide more accurate forecasts with significantly lower variance compared to other models.

%

\begin{table*}[ht]
\vspace{-2mm}
\small
\centering
\caption{Training runtime performance (in seconds).} 
\label{table:training-runtime-results}
\begin{tabular}{@{}l@{} ccccccc@{}} 
\toprule

\textsc{Data} & 
NBEATS & MQRNN & DeepAR & DF & GraphDF-GG & GraphDF-GR & GraphDF-RG \\
\midrule
\textbf{Google}\;\;       &
663.31\ \!$\pm$\ \!54.09  & 284.76\ \!$\pm$\ \!71.08 & 413.79\ \!$\pm$\ \!49.62  & 315.06\ \!$\pm$\ \!67.80  & 279.45\ \!$\pm$\ \!41.19 & \textbf{222.08\ \!$\pm$\ \!69.52} & 281.76\ \!$\pm$\ \!49.51 \\
\textbf{Adobe} &
462.06\ \!$\pm$\ \!120.07 & 393.08\ \!$\pm$\ \!4.22  & 351.99\ \!$\pm$\ \!285.30 & 378.97\ \!$\pm$\ \!441.64 & 282.30\ \!$\pm$\ \!36.80 & \textbf{211.20\ \!$\pm$\ \!21.56} & 264.00\ \!$\pm$\ \!56.29 \\
\bottomrule
\end{tabular}
\vspace{-2mm}
\end{table*}

\begin{table*}[ht]
\vspace{-2mm}
\small
\centering
\caption{Inference runtime performance (in seconds).
}
\label{table:inference-runtime-results}
\begin{tabular}{l ccccccc}
\toprule
\textsc{Data} &
NBEATS & MQRNN & DeepAR & DF & GraphDF-GG & GraphDF-GR & GraphDF-RG \\
\midrule
\textbf{Google} &
88.08\ \!$\pm$\ \!10.96 & 9.22\ \!$\pm$\ \!0.06    & 17.06\ \!$\pm$\ \!0.16    & 8.28\ \!$\pm$\ \!0.02     & 1.67\ \!$\pm$\ \!0.03    & \textbf{0.99\ \!$\pm$\ \!0.003}   & 1.16\ \!$\pm$\ \!0.003 \\
\textbf{Adobe} &
162.63\ \!$\pm$\ \!7.59 & 2.69\ \!$\pm$\ \!0.006   & 4.30\ \!$\pm$\ \!0.02     & 2.12\ \!$\pm$\ \!0.001    & 0.51\ \!$\pm$\ \!0.005   & \textbf{0.28\ \!$\pm$\ \!0.001}   & 0.33\ \!$\pm$\ \!0.000 \\
\bottomrule
\end{tabular}
\end{table*}
%

\subsection{Runtime Analysis} \label{sec:exp-runtime}
We analyze the runtime with respect to both training time and inference time.
Regarding model training time, GraphDF methods are significantly faster and more scalable than other methods, as shown in Table~\ref{table:training-runtime-results}.
In particular, the GraphDF model that uses graph-based global model with RNN-based local model is trained faster than DF, which uses the same RNN-based local model, but differs in the global model used.
This is due to the fact that in the state-of-the-art DF model, all time-series are considered equivalently and jointly when learning the $K$ global factors.
This can be thought of as a fully connected graph where each time-series is connected to every other time-series.
In comparison, the relational global component of GraphDF-GR leverage the graph that encodes explicit dependencies between different time-series, and therefore, does not need to leverage all pairwise time-series, but only a smaller fraction of those that are actually related.
In terms of inference, all models are fast taking only a few seconds as shown in Table~\ref{table:inference-runtime-results}.
For inference, we report the time (in seconds) to infer values in six steps ahead.
In all cases, the GraphDF models are significantly faster than DF and other methods across both Google and Adobe workload datasets.

\subsection{Scalability}
\begin{figure}[ht!]
\centering
\includegraphics[width=0.85\linewidth]{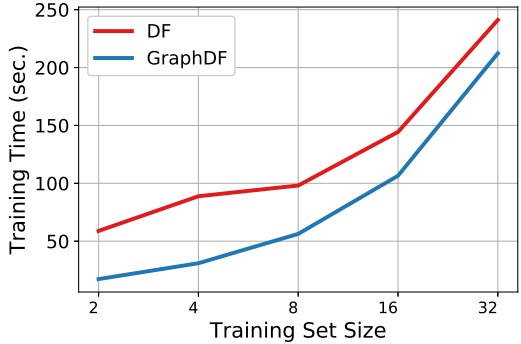}
\vspace{-2mm}
\caption{
Comparing scalability of GraphDF to DF as the training set size increases for the Adobe workload trace dataset.
Note training set size = data points per time-series.
}
\label{fig:training-time-vs-training-set-size}
\vspace{-1mm}
\end{figure}

To evaluate the scalability of GraphDF, we vary the training set size (\ie, the number of previous data points per time-series to use) from $\{2,4,8,16,32\}$, and record the training time.  Fig.~\ref{fig:training-time-vs-training-set-size}  shows that
GraphDF scales nearly linear as the training set size increases from 2 to 32.
For instance, GraphDF takes around 15 seconds to train using only 2 data points per time-series and 30 seconds using 4, and so on.
We also observe that GraphDF is always about 3x faster compared to DF across all training set sizes.

\subsection{$\!\!\!\!\text{Case Study: Opportunistic Scheduling in the Cloud}$}
\label{sec:opportunistic-scheduling}
We leverage our GraphDF forecasting model to enable opportunistic scheduling of batch workloads for cloud resource optimization. Our model generates probabilistic CPU usage forecasts on compute machines, and we use them to schedule batch workloads on machines with low predicted CPU usage. Since batch workloads such as ML training, crawling web pages etc. have loose latency requirements, they can be scheduled on
underutilized resources (such as CPU cores). This improves resource efficiency of the cluster and reduces operating costs by precluding the need to allocate additional machines to run the batch workloads.

The model satisfies following requirements of the scheduling problem. First, the model must be able to correctly forecast utilization. 
If the utilization is underestimated, tasks will be assigned to busy machines and then need to be cancelled, which is a waste of resources.
Second, the execution time of the forecasting model must be significantly faster than the time period used for data collection, \eg, since CPU usage in Google dataset is observed every 5 minutes, the CPU usage forecast should be generated in less than 5 minutes i.e. before the next observation arrives.

We simulate opportunistic scheduling by developing two main components, the \textit{forecaster} and the \textit{scheduler}. The Google dataset simulates the CPU usages for the cluster in this study.
The forecaster reads the 6 most recent observed CPU utilization values of each machine from the data stream and predicts next 3 values. 
The scheduler identifies underutilized machines as those with mean predicted utilization across the three predictions lower than a predefined threshold $\epsilon$ (25\%). Each machine in the cluster has 8 cores and the batch workload requires 6 cores to execute.
To safely make use of the idle resources without disturbing already running tasks or cause thrashing, the scheduler only assigns workloads that require at most (75\%) of compute resources.  
If a machine is assigned batch workloads that exceeds resource availability, they are \emph{terminated/cancelled}.
This procedure is described in Algorithm~\ref{alg:dynamic-real-time-scheduling} in  Appendix~\ref{appendix-opportunistic-learning}.

%
\begin{figure}[t]
\centering
\includegraphics[width=1.0\linewidth]{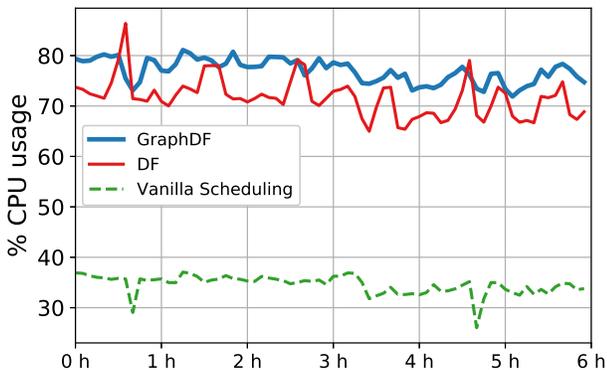}
\vspace{-4mm}
\caption{
    CPU utilization without opportunistic workload scheduling (shown in green) and with scheduling based on each forecaster (shown in red and blue), over a period of 6 hours on Google dataset.
    GraphDF-based scheduling leads to higher CPU utilization than DF-based and vanilla (no forecasts) scheduling.}
\label{fig:cmp_improvement_CPU_utilization-Google-6h}
\vspace{-1mm}
\end{figure}

\begin{table}[b!]
\vspace{-0mm}
	\centering
	\caption{
   Results for opportunistic workload scheduling in the cloud over a 6 hour period using different forecasting models.
	}
	\vspace{1mm}
	\small
	\footnotesize
	\begin{tabular}{@{}cc@{}ccc@{}}
		\toprule
		\multirow{2}{*}{Data} & \multirow{2}{*}{Model}\; & utilization      & correct    & cancellation \\
		                      &                          & improvement (\%) & ratio (\%) & ratio (\%)\\
		\midrule
\TTT\BBB

\multirow{2}{*}{\textbf{Google}} 
& DF      & 38.8           & 68.6          & 20.9          \\
& GraphDF & \textbf{41.9}  & \textbf{88.6} & \textbf{8.2}  \\
\midrule

\multirow{2}{*}{\textbf{Adobe}} 
& DF      & 42.0           & 65.8          & 19.1          \\
& GraphDF & \textbf{53.2}  & \textbf{97.0} & \textbf{2.2}  \\
		\bottomrule
	\end{tabular}
	\label{table:scheduler_metric}
	\vspace{-0mm}
\end{table}


\textbf{Effects on CPU utilization} 
Fig.~\ref{fig:cmp_improvement_CPU_utilization-Google-6h} shows CPU utilization without opportunistic workload scheduling \textit{(vanilla)} and with scheduling based on each forecaster over a period of 6 hours on the Google dataset.
We observe that the GraphDF-based forecaster consistently outperforms both vanilla and DF-based versions by generating forecasts with higher accuracy. 

Table~\ref{table:scheduler_metric} summarizes the performance of the GraphDF-based forecaster with respect to three metrics \emph{CPU utilization improvement}, 
\emph{correct scheduling ratio}, and 
\emph{cancellation ratio}.
The \text{utilization improvement} measures the absolute increase in CPU usage
compared to the vanilla version.
\text{Correct scheduling ratio} corresponds to the ratio when predicted utilization by the scheduler matches the actual utilization.
\text{Cancellation ratio} measures the fraction of machines on which the batch workload was terminated due to incorrectly generated forecasts.
We observe that GraphDF-based  workload scheduling leads to higher CPU utilization, higher correct scheduling ratio, and lower cancellation ratio compared to DF-based scheduling. 
We have included results over longer periods (12 hours and 24 hours) for both datasets in the Appendix~\ref{appendix-additional-results}.

\begin{figure}[!t]
\centering
\includegraphics[width=0.94\linewidth]{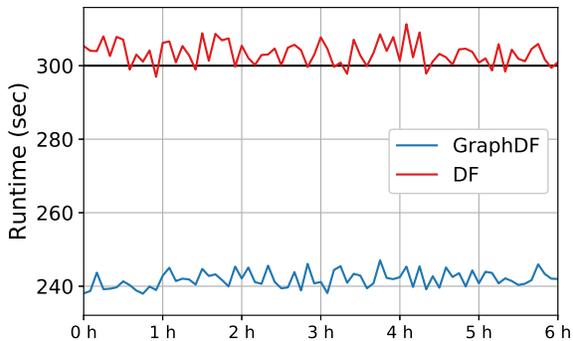}
\vspace{-2mm}
\caption{
    The time constraint (black line), the runtime of scheduler with DF (red line) and that with GraphDF (blue line).
    Note that in most cases, DF fails to meet the time constraint while GraphDF produces a forecast much faster.
}
\vspace{-1mm}
\label{fig:dynamic-runtime}
\end{figure}
\textbf{Execution Time Comparison} 
Fig.~\ref{fig:dynamic-runtime} shows that the runtime of DF-based scheduling often
exceeds the 5-minute time limit, while GraphDF-based version is much faster and always meets it. Hence, GraphDF persuasively provides a solution for enhancing cloud efficiency through effective usage forecasting.

\section{Conclusion} \label{sec:conc}
In this work, we introduced a deep hybrid graph-based probabilistic forecasting framework called Graph Deep Factors.
While existing deep probabilistic forecasting approaches do not explicitly leverage a graph, and assume either complete independence among time-series (\ie, completely disconnected graph) or complete dependence between all time-series (\ie, fully connected graph), this work moved beyond these two extreme cases by 
allowing nodes and their time-series to have arbitrary and explicit weighted dependencies among each other.
Such explicit and arbitrary weighted dependencies between nodes and their time-series are modeled as a graph in the proposed framework.
Notably, GraphDF consists of a relational global model that learns complex non-linear time-series patterns globally using the structure of the graph to improve computational efficiency as well as a relational local model that not only considers its individual time-series but the time-series of nodes that are connected in the graph to improve forecasting accuracy.
Finally, the experiments demonstrated the effectiveness of the proposed deep hybrid graph-based probabilistic forecasting model in terms of its forecasting performance, runtime, scalability, 
and optimizing cloud usage through opportunistic workload scheduling.

\clearpage
\newpage

\bibliography{main}

\bibliographystyle{mlsys2020}

\clearpage
\newpage

\appendix 
\section*{Appendix}
\label{sec:appendix}

\section{Alternative Models} \label{appendix-alternative-models}

\subsection{Learning Relational Global Factors via DCRNN}
\label{appendix:dcrnn}
For the relational global component of GraphDF, we can also leverage DCRNN~\cite{dcrnn18}.
Different from the GCRN model, the original DCRNN leverages a diffusion convolution operation and a GRU structure for learning the relational global factors of GraphDF.

Given the time-series graph signal $\mY_t \in \RR^{N\times P}$ with $N$ nodes, the diffusion convolution with respect to the graph-based time-series is defined as,
\begin{align}
    f_\gconv{\mTheta}(\mY_t) = \sum_{l=0}^{L-1}(\theta_{l}\tmA^l)\mY_t
\end{align}
where $\tmA=\mD^{-1}\mA$ is the normalized adjacency matrix of the graph $G$ that captures the explicit weighted dependencies between the multi-dimensional time-series of the nodes.
The Chebyshev polynomial approximation is used similarly as Eq.~\eqref{eq:filt_cheby}.

The relational global factors are learned using the graph diffusion convolution combined with GRU enabling them to be carried forward over time using the graph structure,
\begin{align}
\mR_t &= \sigma(\mTheta_R \gconv{}[\mY_t,\; \mH_{t-1}] + \vb_R) \\
\mU_t &= \sigma(\mTheta_U \gconv{}[\mY_t,\; \mH_{t-1}] + \vb_U) \\
\mC_t &= \tanh(\mTheta_C \gconv{}[\mY_t,\; (\mR_t \odot \mH_{t-1})] + \vb_C) \\
\mH_t &= \mU_t \odot \mH_{t-1} + (1-\mU_t) \odot \mC_t \label{eq:dcrnn-ht}
\end{align}
where $\mH_t \in \RR^{N\times Q}$ denotes the hidden state of the model at time step $t$,
$Q$ is the number of hidden units,
$\mR_t \in \RR^{N\times Q}, \mU_t \in \RR^{N\times Q}$ are called as reset gate and update gate at time $t$, respectively.
$\mTheta_R \in \RR^{L}, \mTheta_U \in \RR^{L}, \mTheta_C \in \RR^{L}$ denote the parameters corresponding to different filters.

With the hidden state $\mH_t$ in Eq.~\eqref{eq:dcrnn-ht}, the fixed effect is derived from DCRNN similarly with Eq.~\eqref{eq:gcrn-st} and Eq.~\eqref{eq:gcrn-cit}.
Compared to the previous GCRN that we adapted for the relational global component, DCRNN is more computationally efficient due to the GRU structure it uses.

\subsection{Relational Local Model via Probabilistic DCRNN}
\label{sec:prob-DCRNN}
We also describe a probabilistic DCRNN for the relational local component of GraphDF.
For a given node $i$, its relational local random effect is derived with respect to its past observations, covariates and those of its neighbors, denoted by $\mY_t^{(i)} \in \RR^{(|S|+1) \times P}$ as defined in Eq.~\eqref{eq:Y_t_rel-local}.
The diffusion convolution models the relational local random effect among nodes.
The GRU structure is adapted with the diffusion convolution to allow the random effects to be forwarded in time.
\begin{align}
    \mR_t^{(i)} &= 
    \sigma\big(
    \mTheta_{R}^{(i)} \gconv{}
    [\mY_t^{(i)},\mH_t^{(i)}] + \vb_R^{(i)}
    \big)\\
    \mU_t^{(i)} &= 
    \sigma\big(
    \mTheta_{U}^{(i)} \gconv{}
    [\mY_t^{(i)},\mH_t^{(i)}] + \vb_U^{(i)}
    \big)\\
    \mC_t^{(i)} &= 
    \tanh\big(
    \mTheta_{C}^{(i)} \gconv{}
    [\mY_t^{(i)},\mH_t^{(i)}] + \vb_C^{(i)}
    \big)\\
    \mH_t^{(i)} &= 
    \mU_t^{(i)}\odot\mH_{t-1}^{(i)} +
    (1 - \mU_t^{(i)})\odot\mC_{t}^{(i)}
\end{align}
where $\mTheta_{R}^{(i)} \in \RR^{P\times R}, \mTheta_{U}^{(i)} \in \RR^{P\times R}, \mTheta_{C}^{(i)} \in \RR^{P\times R}$ denote the parameters corresponding to different filters,
$\mH_t^{(i)} \in \RR^{(|C|+1)\times R}$ is the hidden state for node $i$ and its neighbors $\Gamma_i$,
$R$ is the number of hidden units in the relational local model.
$\vb_{I}^{(i)}, \vb_{F}^{(i)}, \vb_{C}^{(i)}, \vb_{O}^{(i)}$ are bias vector parameters.

The graph convolution in equations above is performed with the submatrix $\mL^{(i)}$ taken from the Laplacian matrix $\mL$ of the graph $G$ that explicitly models the important and meaningful dependencies between the multi-dimensional time-series data of each node.
The matrix $\mL^{(i)}$ consists of rows and columns corresponding to node $i$ and its neighbors $\Gamma_i$.

With the hidden state $\mH_t^{(i)}$, the relational local random effect $b_t^{(i)}(\cdot)$ is calculated similarly with Eq.~\eqref{eq:sigma-gcrn} and Eq.~\eqref{eq:gcrn-random}.

\begin{table*}[!t]
\small
\centering
\setlength{\tabcolsep}{4.6pt}
\caption{Results for one-step ahead forecasting (\textsc{p10ql}).
}
\vspace{1mm}
\label{table:results-one-step-p10ql}
\begin{tabular}{
l ccccccc
}
\toprule
\multirow{1}{*}{\textsc{data}}       &
NBEATS & MQRNN & DeepAR & DF & GraphDF-GG & GraphDF-GR & GraphDF-RG \\
\midrule
\TTT\BBB
\multirow{1}{*}{\textbf{Google}\;\;} &
18.116\ \!$\pm$\ \!201.259           & 0.190\ \!$\pm$\ \!0.004 & 0.046\ \!$\pm$\ \!0.000 & 0.083\ \!$\pm$\ \!0.001 & \textbf{0.037\ \!$\pm$\ \!0.000} & 0.038\ \!$\pm$\ \!0.000 & 0.044\ \!$\pm$\ \!0.000 \\
\multirow{1}{*}{\textbf{Adobe}}      &
0.615\ \!$\pm$\ \!0.091& 0.132\ \!$\pm$\ \!0.000 & 0.164\ \!$\pm$\ \!0.001 & 1.128\ \!$\pm$\ \!0.004 & \textbf{0.118\ \!$\pm$\ \!0.001} & 0.119\ \!$\pm$\ \!0.000 & 1.027\ \!$\pm$\ \!2.700 \\
\bottomrule
\end{tabular}
\end{table*}

\begin{table*}[!ht]
\setlength{\tabcolsep}{5.2pt} 
\centering
\caption{Results for multi-step ahead forecasting (\textsc{p10ql}).
}
\vspace{1mm}
\label{table:results-multi-step-ahead-forecasting-p10ql}
\small
\begin{tabular}{@{}l@{} cccccccc 
}
\toprule
\multirow{1}{*}{\textsc{data}} 
& \multirow{1}{*}{\textsc{h}} & NBEATS & MQRNN & DeepAR & DF & GraphDF-GG & GraphDF-GR & GraphDF-RG \\
\midrule
\multirow{3}{*}{\textbf{Google}\;\;}
& 3 & 0.652\ \!$\pm$\ \!0.396 & 0.152\ \!$\pm$\ \!0.006 & 0.070\ \!$\pm$\ \!0.000 & 0.132\ \!$\pm$\ \!0.004 & \textbf{0.064\ \!$\pm$\ \!0.001} & 0.077\ \!$\pm$\ \!0.000 & 0.087\ \!$\pm$\ \!0.000 \\
& 4 & 0.260\ \!$\pm$\ \!0.017 & 0.272\ \!$\pm$\ \!0.018 & 0.138\ \!$\pm$\ \!0.000 & 0.193\ \!$\pm$\ \!0.016 & \textbf{0.071\ \!$\pm$\ \!0.001} & 0.083\ \!$\pm$\ \!0.000 & 0.089\ \!$\pm$\ \!0.001 \\
& 5 & 0.447\ \!$\pm$\ \!0.054 & 0.147\ \!$\pm$\ \!0.005 & 0.484\ \!$\pm$\ \!0.017 & 0.327\ \!$\pm$\ \!0.036 & \textbf{0.054\ \!$\pm$\ \!0.000} & 0.113\ \!$\pm$\ \!0.001 & 0.088\ \!$\pm$\ \!0.001 \\

\midrule
\multirow{3}{*}{\textbf{Adobe}}
& 3 & 0.811\ \!$\pm$\ \!0.295 & 0.184\ \!$\pm$\ \!0.003 & 0.207\ \!$\pm$\ \!0.002 & 0.303\ \!$\pm$\ \!0.006 & \textbf{0.183\ \!$\pm$\ \!0.002} & 0.216\ \!$\pm$\ \!0.008 & 0.267\ \!$\pm$\ \!0.009 \\
& 4 & 0.985\ \!$\pm$\ \!0.537 & 0.219\ \!$\pm$\ \!0.008 & 0.273\ \!$\pm$\ \!0.003 & 0.313\ \!$\pm$\ \!0.018 & \textbf{0.184\ \!$\pm$\ \!0.002} & 0.242\ \!$\pm$\ \!0.014 & 0.423\ \!$\pm$\ \!0.019 \\
& 5 & 0.626\ \!$\pm$\ \!0.023 & 0.398\ \!$\pm$\ \!0.229 & 0.402\ \!$\pm$\ \!0.011 & 0.343\ \!$\pm$\ \!0.047 & \textbf{0.251\ \!$\pm$\ \!0.016} & 0.298\ \!$\pm$\ \!0.031 & 0.544\ \!$\pm$\ \!0.020 \\

\bottomrule
\end{tabular}
\end{table*}

\section{Experimental details}
\subsection{Data} \label{appendix-data-details}

\smallskip\noindent\textit{Google Trace.}~\cite{reiss2011google}
The Google trace dataset records the activities of a cluster of $12,580$ machines for 29 days since 19:00 EDT in May 1, 2011.
The CPU and memory usage for each task are recorded every 5 minutes. 
The usage of tasks is aggregated to the usage of associated machines, resulting time-series of length $8,354$.

\medskip\noindent\textit{Adobe Workload Trace.}
The Adobe trace dataset records the CPU and memory usage of $3,270$ nodes in the period from Oct. 31 to Dec. 5 in 2018.
The timescale is 30 minutes, resulting time-series of length $1,687$.

\medskip\noindent\textit{Graph Construction.}
For each dataset, we derive a graph where each node represents a machine with one or more time-series associated with it, and each edge represents the similarity between the node time-series $i$ and $j$. 
The constructed graph encodes the dependency information between nodes.
In this work, we estimate the edge weights using the radial basis function (RBF) kernel with the previous time-series observations as 
$K(\vz_i,\vz_j) = \exp(-\frac{\left\Vert\vz_i-\vz_j\right\Vert^2}{2\ell^2})$, 
where $\ell$ is the length scale of the kernel.

For all experiments, we use a single machine equipped with Linux Ubuntu OS with an 8-core CPU for training and inference.

\subsection{Hyperparameter Tuning} \label{appendix:hyperparameter-tuning}
Most hyperparameters are set as default values by MXnet Gluonts~\cite{Alexandrov2019GluonTSPT}. We list hyperparameters used in our experiments as follows:
\begin{itemize}
    \item learning rate decay factor: $0.5$
    \item minimum learning rate: $5*10^{-5}$
    \item weight initializer method: Xavier initialization
    \item training epochs, we train for 500 epochs when using the Adobe dataset and 100 epochs when using the Google dataset.
\end{itemize}
Some hyperparameters are specific to our method:
In GraphDF, we set the order $L=1$ in Eq.~\eqref{eq:filt_cheby}.
A small number of the order indicates the model make forecasts based more on neighboring nodes than those in distance.
For other methods, we use default hyperparameters given by the Gluonts implementation if not mentioned.

\section{Additional Forecasting Results} \label{appendix-additional-results}
The Quantile Loss at 10th percentile is showed in Table~\ref{table:results-one-step-p10ql} and Table~\ref{table:results-multi-step-ahead-forecasting-p10ql}.
The result shows that our hybrid graph models GraphDF and variants consistently 
outperform previous state-of-the-arts.
Specifically, GraphDF with graph models in both relational global component and relation local component obtain the best performance compared to other methods.

\section{Additional Results for Opportunistic Scheduling}
\label{appendix-opportunistic-learning}
Algorithm~\ref{alg:dynamic-real-time-scheduling} provides an overview of the opportunistic real-time scheduler.
Notice that we can leverage any forecasting model $f$ to obtain forecasts of CPU usage which is then used in the scheduler.

\begin{figure}[!ht]
\centering
\includegraphics[width=0.85\linewidth]{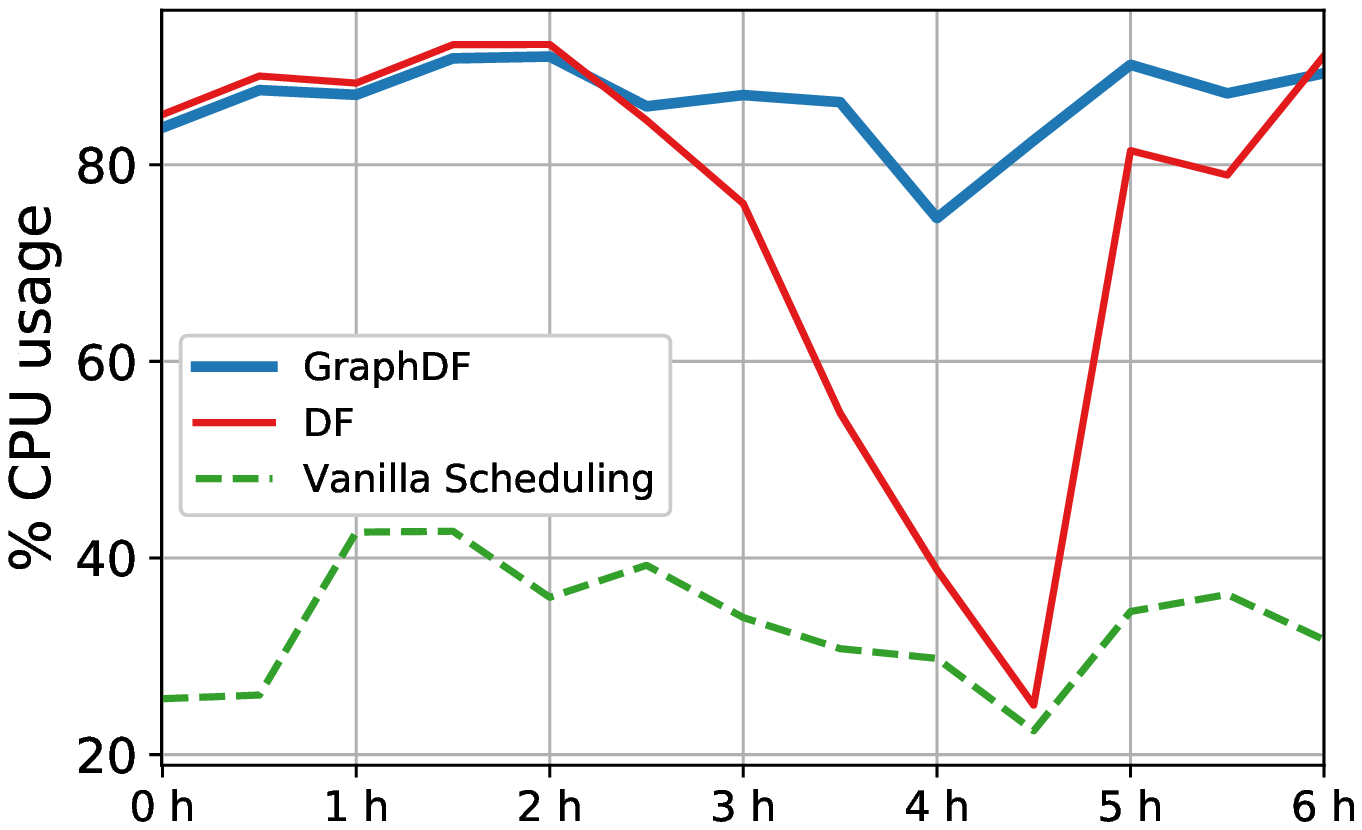}
\includegraphics[width=0.85\linewidth]{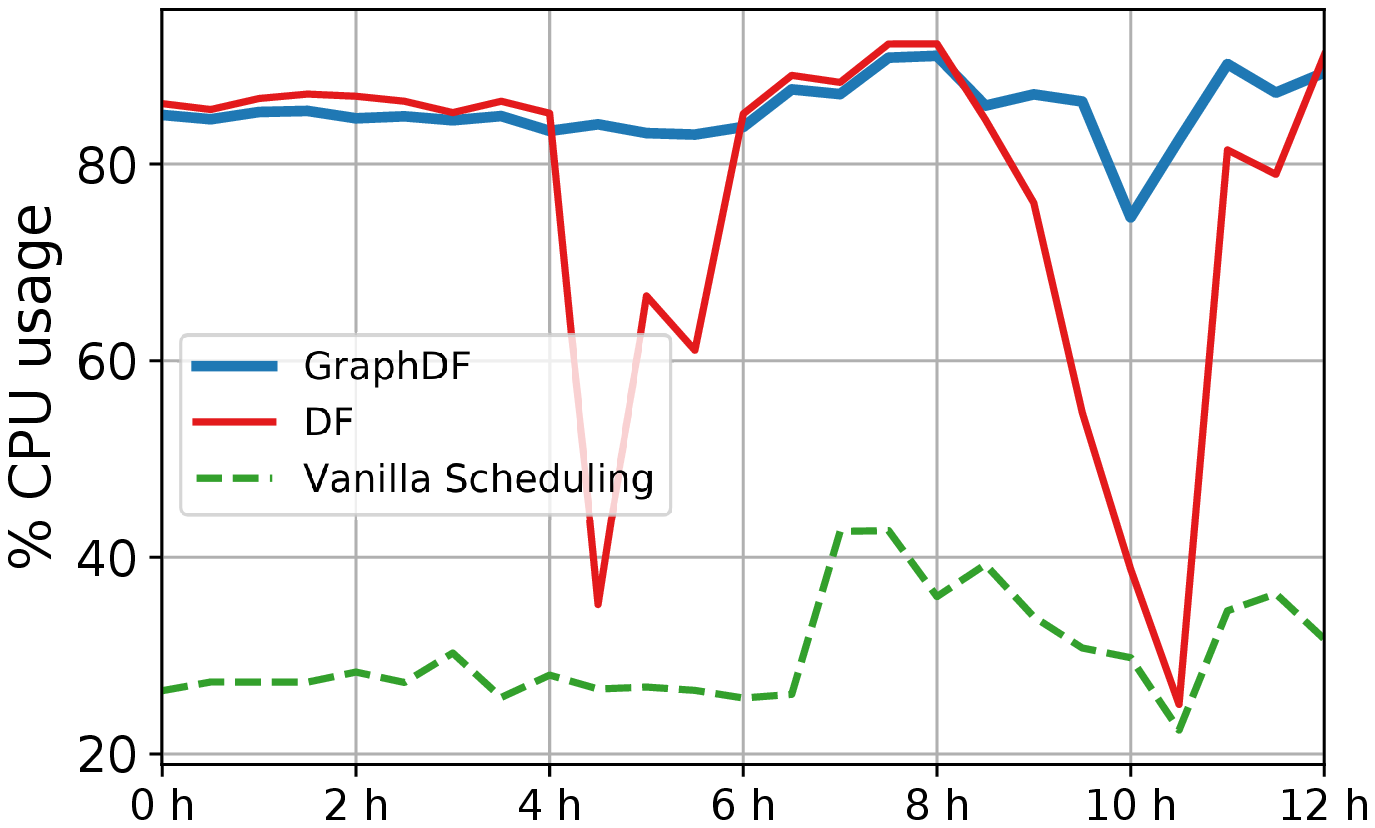}
\includegraphics[width=0.85\linewidth]{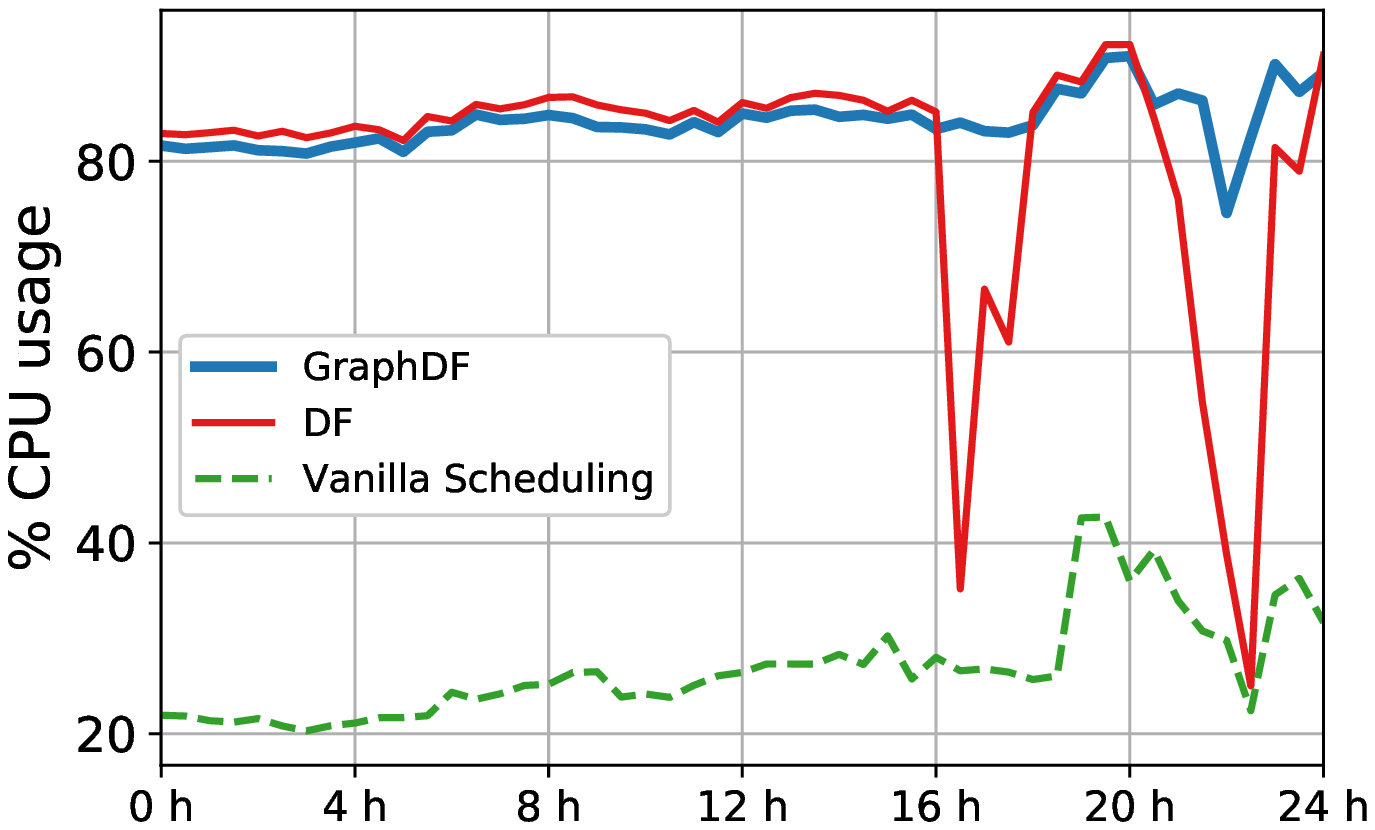}
\caption{
CPU utilization without opportunistic workload scheduling (shown in green) and with scheduling based on each forecaster (shown in red and blue), over a period of 6, 12 and 24 hours on Adobe dataset.
GraphDF-based scheduling leads to higher CPU utilization than DF-based and vanilla (no forecasts) scheduling.
}
\label{fig:cmp_improvement_CPU_utilization-Adobe}
\end{figure}

\begin{figure}[!ht]
\centering
\includegraphics[width=0.9\linewidth]{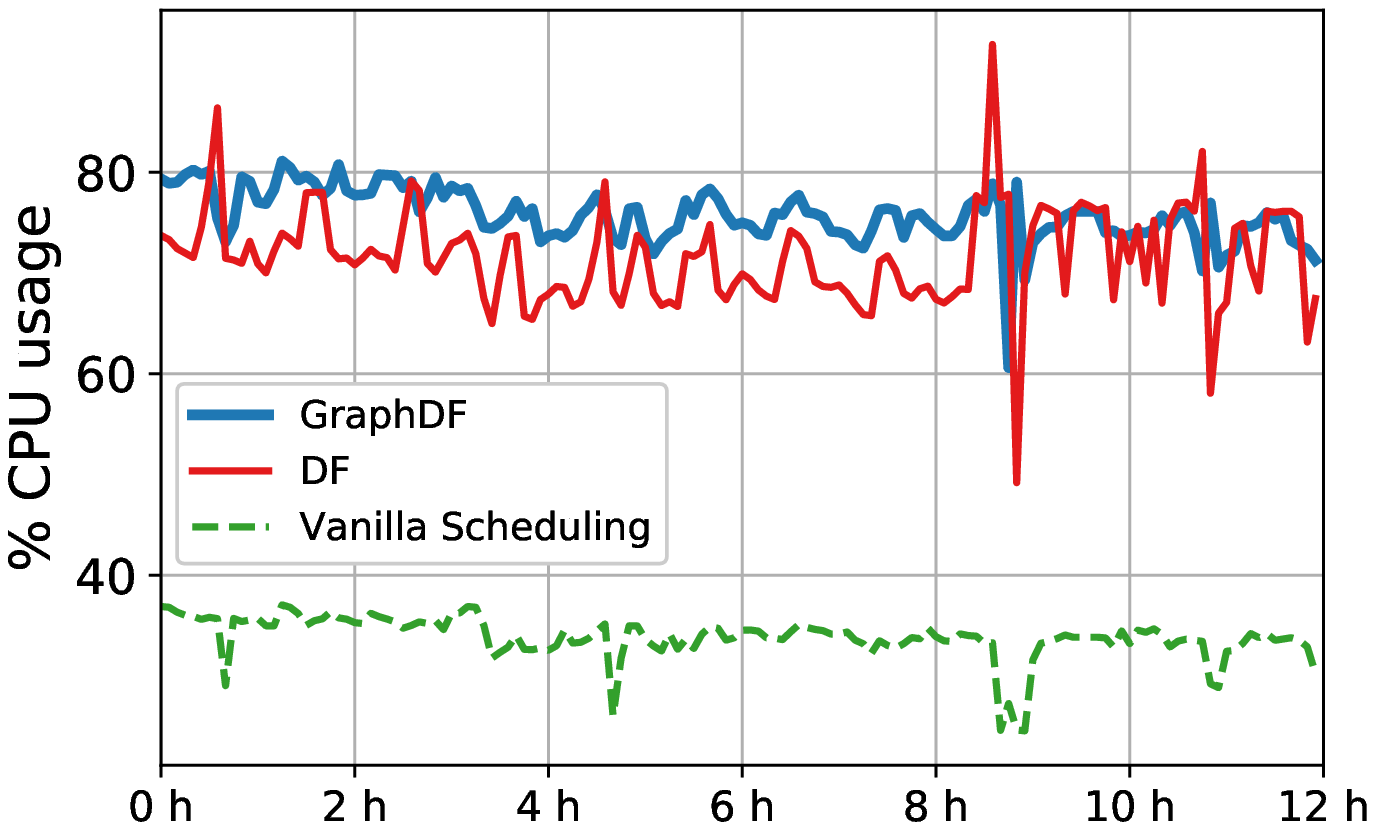}
\includegraphics[width=0.9\linewidth]{graphics/G-improv9.eps}
\caption{
CPU utilization without opportunistic workload scheduling (shown in green) and with scheduling based on each forecaster (shown in red and blue), over a period of 12 and 24 hours on Google dataset.
GraphDF-based scheduling leads to higher CPU utilization than DF-based and vanilla (no forecasts) scheduling.
}
\label{fig:cmp_improvement_CPU_utilization-Google}
\end{figure}

More result on for opportunistic scheduling with different forecasters are depicted in Fig.~\ref{fig:cmp_improvement_CPU_utilization-Adobe} and Fig.~\ref{fig:cmp_improvement_CPU_utilization-Google}.
From the figures, we observe that schedulers with either forecaster improve CPU utilization, whereas scheduler with our proposed GraphDF forecaster perform better in terms of utilization improvement.

\begin{algorithm}[ht!]
    \caption{Opportunistic Scheduling}
    \label{alg:dynamic-real-time-scheduling}
    \begin{algorithmic}[1] 
    \STATE Initialize hyperparameters and variables lookback window $w=6$, horizon $\tau=3$, threshold ratio $\epsilon=25\%$, portion ratio $\lambda=75\%$. Accumulated utilization improvement $Acc=0$
    \WHILE{New observations arrive $t\gets t+1$}
        \STATE Initialize weights $\mPhi$ for new model $f_t$
        \STATE $f_t \gets \mathbb{P}\big(\cdot \big| \mPhi, \mA, \big\{\mX_{:,t-w+1:t+\tau}^{(i)}, \vz_{t-w+1:t}^{(i)}\big\}_{i=1}^{N} \!\big)$
        \FOR{each node $i$}
            \STATE Obtain forecasts
            $\hat{z}^{(i)}\!\!=\!\!\{\hat{z}_{t+1}^{(i)}, \hat{z}_{t+2}^{(i)}\ldots\hat{z}_{t+\tau}^{(i)}\}\!\!\sim\!\! f_t$
            \IF {MEAN of forecasts MEAN$(\hat{z}^{(i)}) \leq \epsilon$}
                \STATE Utilization $Acc\!\gets\!\!Acc+\lambda(1-\text{MEAN}(\hat{z}^{(i)}))$
            \ELSE 
                \STATE Cancel assigned tasks on node $i$
            \ENDIF
        \ENDFOR
    \ENDWHILE
    \end{algorithmic}       
\end{algorithm}

\begin{table}[bt!]
\setlength{\tabcolsep}{2.5pt}
	\centering
	\caption{Opportunistic scheduling performance using different forecasting models.
	}
	\vspace{0mm}
	\small
	\footnotesize
	\begin{tabular}{lccccc}
		\toprule
		\multirow{2}{*}{Data} & \multirow{2}{*}{Hour} & \multirow{2}{*}{Model} & utilization & correct    & cancellation \\
        & & & improvement (\%)\; & ratio (\%) & ratio (\%)\\
		\midrule
\TTT\BBB
\multirow{2}{*}{\textbf{G}} & \multirow{2}{*}{\textbf{12}}
& DF & 37.8  & 60.0 & 31.8 \\
& & GraphDF & \textbf{41.9}  & \textbf{87.6} & \textbf{9.2}  \\
\midrule
\multirow{2}{*}{\textbf{A}} & \multirow{2}{*}{\textbf{12}}
& DF & 53.6  & 76.9 & 17.6 \\
& & GraphDF & \textbf{57.1}  & \textbf{97.4} & \textbf{1.8}  \\
\midrule
\multirow{2}{*}{\textbf{G}} & \multirow{2}{*}{\textbf{24}}
& DF & 42.6  & 51.2 & 43.6 \\
& & GraphDF & \textbf{43.8}  & \textbf{85.4} & \textbf{10.0} \\
\midrule
\multirow{2}{*}{\textbf{A}} & \multirow{2}{*}{\textbf{24}}
& DF & 59.6  & 78.6 & 17.1 \\
& & GraphDF & \textbf{62.5}  & \textbf{98.1} & \textbf{1.2}  \\
		\bottomrule
	\end{tabular}
	\label{table:scheduler_metric_12_24hours}
\end{table}

\section{Evaluation of Probabilistic Forecasts} \label{appendix:prob-forecast-eval-metric}
To evaluate the probabilistic forecasts, we use the quantile loss defined as follows:
given a quantile $\rho\in(0,1)$, a target value $\vz_t$ and $\rho$-quantile prediction $\widehat{\vz}_t(\rho)$, the $\rho$-quantile loss is defined as
\begin{align}
\text{QL}_\rho[\vz_t, \widehat{\vz}_t(\rho)] &= 2\big[\rho(\vz_t - \widehat{\vz}_t(\rho))\mathbb{I}_{\vz_t - \widehat{\vz}_t(\rho) > 0} \nonumber \\
&+ (1-\rho)(\widehat{\vz}_t(\rho) - \vz_t)\mathbb{I}_{\vz_t - \widehat{\vz}_t(\rho) \leqslant 0}\big]
\end{align}
For deriving quantile losses over a time-span across all time-series, we adapt to a normalized version of quantile loss $\sum_{i,t} \text{QL}_\rho[z_{i,t}, \hat{z}_{i,t}(\rho)] / \sum_{i,t} |z_{i,t}|$.
We run 10 trials and report the average 
for $\rho=\{0.1, 0.5, 0.9\}$, denoted as the P10QL, P50QL and P90QL, respectively.
In experiments, the quantile losses are computed based on 100 sample values.

\subsection{An overview of the model training} \label{appdendix-model-training}
We summarize the training procedure in Algorithm~\ref{alg:graphDF-training}.
\begin{algorithm}[h!]
    \caption{Training Graph Deep Factor Models}
    \label{alg:graphDF-training}
    \begin{algorithmic}[1] 
    \STATE Initialize the parameters $\mPhi$
    \FOR{each time series pair $\{(\vz^{(i)}, \vx^{(i)})\}$}
        \STATE 
        Using current model parameter estimates $\mPhi$, derive the relational fixed effect as
        $
        c_{t}^{(i)} = \sum_{k=1}^K w_{i,k}\cdot S_{i,k,t}
        $ and relational local random effect $b_{t}^{(i)}$.
      \STATE Compute marginal likelihood $\mathbb{P}(\vz^{(i)})$ and accumulate the loss.
    \ENDFOR
    \STATE Compute the overall loss in the current batch and perform 
    stochastic gradient descent and update the trainable parameters $\mPhi$ accordingly.
    \end{algorithmic}       
\end{algorithm}

\end{document}